\documentclass[sn-mathphys,iicol]{sn-jnl}

\usepackage[utf8]{inputenc} 
\usepackage[T1]{fontenc}    
\usepackage{booktabs}       
\usepackage{amsfonts}       
\usepackage{nicefrac}       
\usepackage{microtype}      
\usepackage{eqparbox}
\usepackage{tablefootnote}
\usepackage{stackengine}
\usepackage{enumitem}
\usepackage{url}
\PassOptionsToPackage{obeyspaces}{url}
\usepackage{amsmath}
\usepackage{multirow}
\usepackage{graphicx}
\usepackage{wrapfig}
\usepackage{threeparttable}
\usepackage{subfigure}
\usepackage{amsmath}
\usepackage{verbatim}
\usepackage{xcolor}

\jyear{2021}%

\theoremstyle{thmstyleone}%
\theoremstyle{thmstyletwo}%

\theoremstyle{thmstylethree}%

\begin{document}

\title[Article Title]{DGL-GAN: Discriminator Guided GAN Compression}


\author[1]{\fnm{Yuesong} \sur{Tian}}\email{tianys163@gmail.com}

\author[2]{\fnm{Li} \sur{Shen}}\email{mathshenli@gmail.com}

\author*[1,3]{\fnm{Xiang} \sur{Tian}}\email{tianx@zju.edu.cn}

\author[2]{\fnm{Dacheng} \sur{Tao}}\email{dacheng.tao@gmail.com}

\author*[4]{\fnm{Zhifeng} \sur{Li}}\email{michaelzfli@tencent.com}

\author*[4]{\fnm{Wei} \sur{Liu}}\email{wl2223@columbia.edu}

\author[1,5]{\fnm{Yaowu} \sur{Chen}}\email{cyw@mail.bme.zju.edu.cn}

\affil*[1]{\orgdiv{College of Biomedical Engineering and Instrument
Science}, \orgname{Zhejiang University}, \orgaddress{\city{Hangzhou}, \country{China}}}

\affil[2]{\orgname{JD Explore Academy}, \orgaddress{\city{Beijing}, \country{China}}}

\affil[3]{\orgdiv{Zhejiang Provincial Key Laboratory for Network Multimedia Technologies}, \orgname{Zhejiang University}, \orgaddress{\city{Hangzhou}, \country{China}}}

\affil[4]{\orgname{Tencent}, \orgaddress{\city{Shenzhen}, \country{China}}}

\affil[5]{\orgdiv{Zhejiang University Embedded System Engineering Research Center}, \orgname{Zhejiang University}, \orgaddress{\city{Hangzhou}, \country{China}}}


\abstract{Generative Adversarial Networks (GANs) with high computation costs, e.g., BigGAN and StyleGAN2, have achieved remarkable results in synthesizing high-resolution images from random noise. Reducing the computation cost of GANs while keeping generating photo-realistic images is a challenging field. In this work, we propose a novel yet simple {\bf D}iscriminator {\bf G}uided {\bf L}earning approach for compressing vanilla {\bf GAN}, dubbed {\bf DGL-GAN}. Motivated by the phenomenon that the teacher discriminator may contain some meaningful information about both real images and fake images, we merely transfer the knowledge from the teacher discriminator via the adversarial interaction between the teacher discriminator and the student generator. We apply DGL-GAN to compress the two most representative large-scale vanilla GANs, i.e., StyleGAN2 and BigGAN. Experiments show that DGL-GAN achieves state-of-the-art (SOTA) results on both StyleGAN2 and BigGAN. Moreover, DGL-GAN is also effective in boosting the performance of original uncompressed GANs. Original uncompressed StyleGAN2 boosted with DGL-GAN achieves FID 2.65 on FFHQ, which achieves a new state-of-the-art performance. Code and models are available at \url{https://github.com/yuesongtian/DGL-GAN}}




\maketitle
 
\clearpage
\section{Introduction}\label{sec:introduction}

GANs firstly proposed in~\cite{goodfellow2014generative} have been prevalent for computer vision tasks, see, e.g., image generation \cite{karras2019style, karras2020analyzing, brock2018large, sauer2022stylegan, sauer2023stylegan, kang2023scaling}, image editing \cite{shen2020interpreting, kim2021stylemapgan, li2021surrogate, lin2021anycost, zhuang2021enjoy}, image denoising \cite{kaneko2020noise, dutta20193d}, and the references therein. 
For the noise-to-image generation, two of the most representative vanilla GANs are StyleGAN series \cite{karras2019style, karras2020analyzing, karras2021alias} and BigGAN \cite{brock2018large}, which can generate diverse photo-realistic images from random noise by using a huge number of parameters and large scale high-quality training datasets. Roughly speaking, StyleGAN2 contains $30.37$M parameters on Flickr-Faces-HQ (FFHQ) \cite{karras2019style}, and BigGAN contains $70.33$M parameters on ImageNet \cite{deng2009imagenet}. Deploying such large-scale GANs on computational resource-limited devices is an urgent and challenging task, which largely impedes its broader applications.  

\begin{figure*}[h!]
  \centering
  \subfigure[\texttt{FFHQ}]{\includegraphics[width=0.48\textwidth]{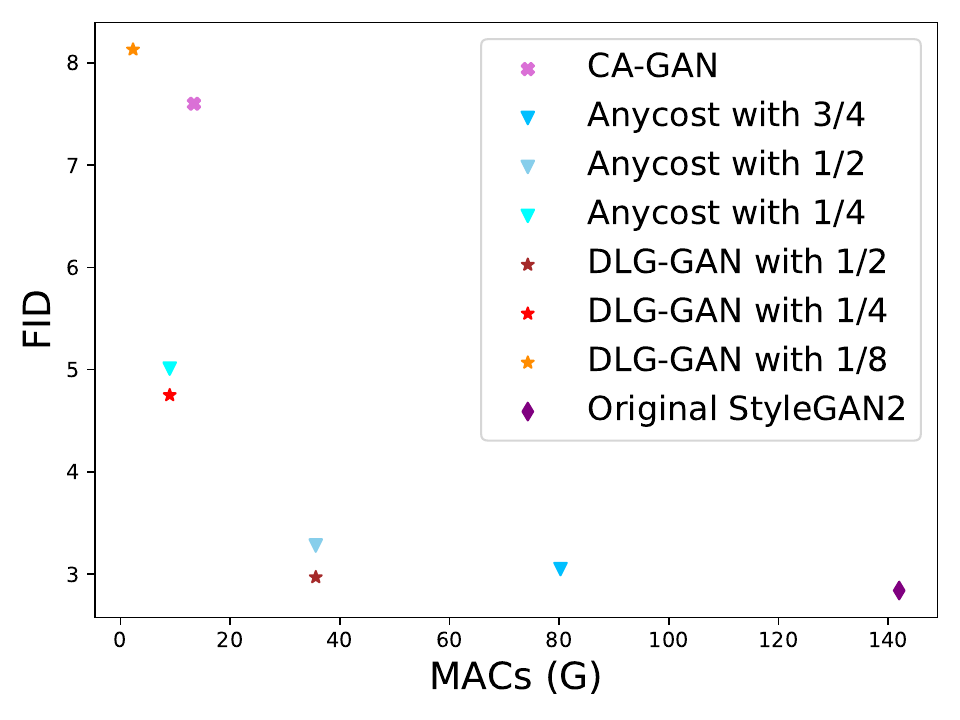}}
  \subfigure[\texttt{ImageNet}]{\includegraphics[width=0.48\textwidth]{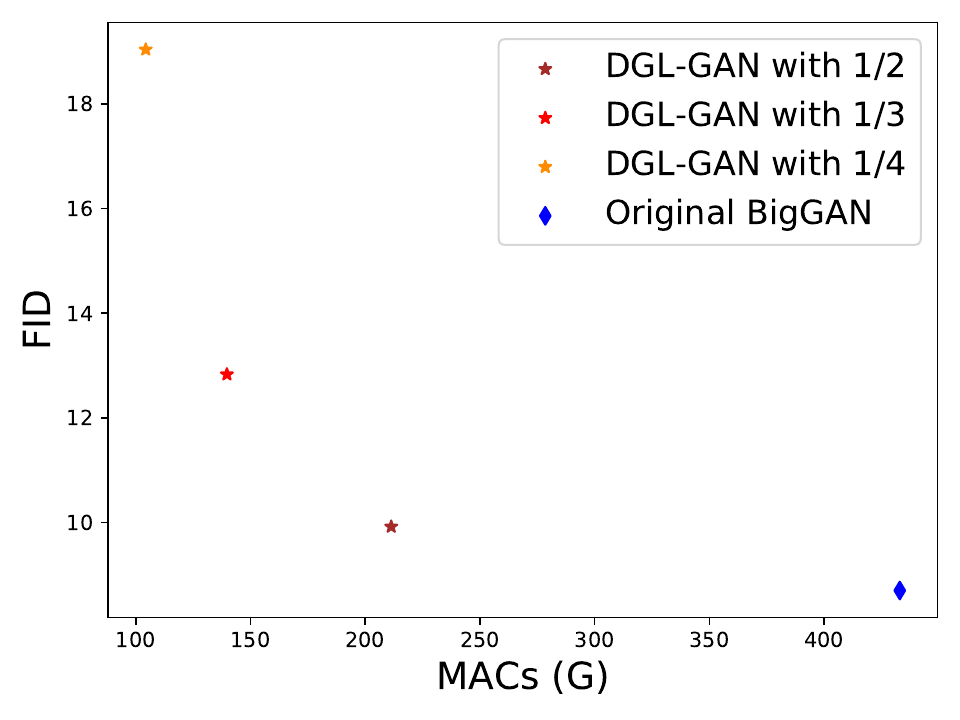}}
  \vspace{-0.3cm}
  \caption{\small The MACs-Performance curve on FFHQ and ImageNet. (a) is the Params-Performance curve on compressing StyleGAN2 \cite{karras2020analyzing}. (b) is the MACs-Performance curves on compressing BigGAN \cite{brock2018large}.}
  \label{Fig: Performance_curve}
\end{figure*}

To make GANs easy to be deployed on computational resource limited devices, extensive works have been proposed to obtain lightweight GANs. The mainstreaming approach is to inherit the model compression techniques developed for image-classification task to compress GANs, such as weight pruning \cite{chen2021gans}, weight quantization \cite{wang2019qgan}, channel pruning \cite{wang2020gan,liu2021content,hou2020slimmable,shen2020cpot}, lightweight GAN architecture search/design \cite{tian2020alphagan,liu2021towards,belousov2021mobilestylegan,wang2021coarse}, evolutionary compression \cite{li2021evolutionary,wang2019evolutionary}, and knowledge distillation (KD) \cite{aguinaldo2019compressing,yu2020self,chen2020distilling,li2020gan,jin2021teachers,li2020learning,fu2020autogan,zhang2018self}. 
However, most of the above works focus on compressing conditional (cycle) GANs for image-to-image generation tasks, scarce works have been proposed for compressing vanilla GANs except recent works \cite{liu2021content, hou2020slimmable, yu2020self, lin2021anycost, li2021revisiting, kang2022information}. The difficulties of compressing vanilla GANs mainly lie in four aspects: 
\begin{enumerate}[label=(\alph*)]
\item \emph{The existing pretrained generator is sensitive to the reduction of parameters.}\label{difficulty-a}
Several compression methods \cite{liu2021content, lin2021anycost} conduct pruning on the pretrained generator and further tune it. However, pruning the pretrained generator will inevitably degrade the quality of synthesized images, because every pixel in the synthesized image relates to all channels of convolutional kernels or intermediate features in the generator, also pointed out in \cite{yu2020self}. To illustrate that, we mask $1/8$ parameters of the convolutional kernels in the pretrained StyleGAN2 \cite{karras2020analyzing} and generate images with it, presented in Fig. \ref{Fig: Mask Channels StyleGAN2}, where the quality of images is  intensely affected.

\begin{figure}[h]
  \centering
  \subfigure[\texttt{Mask}]{\includegraphics[width=1.8cm]{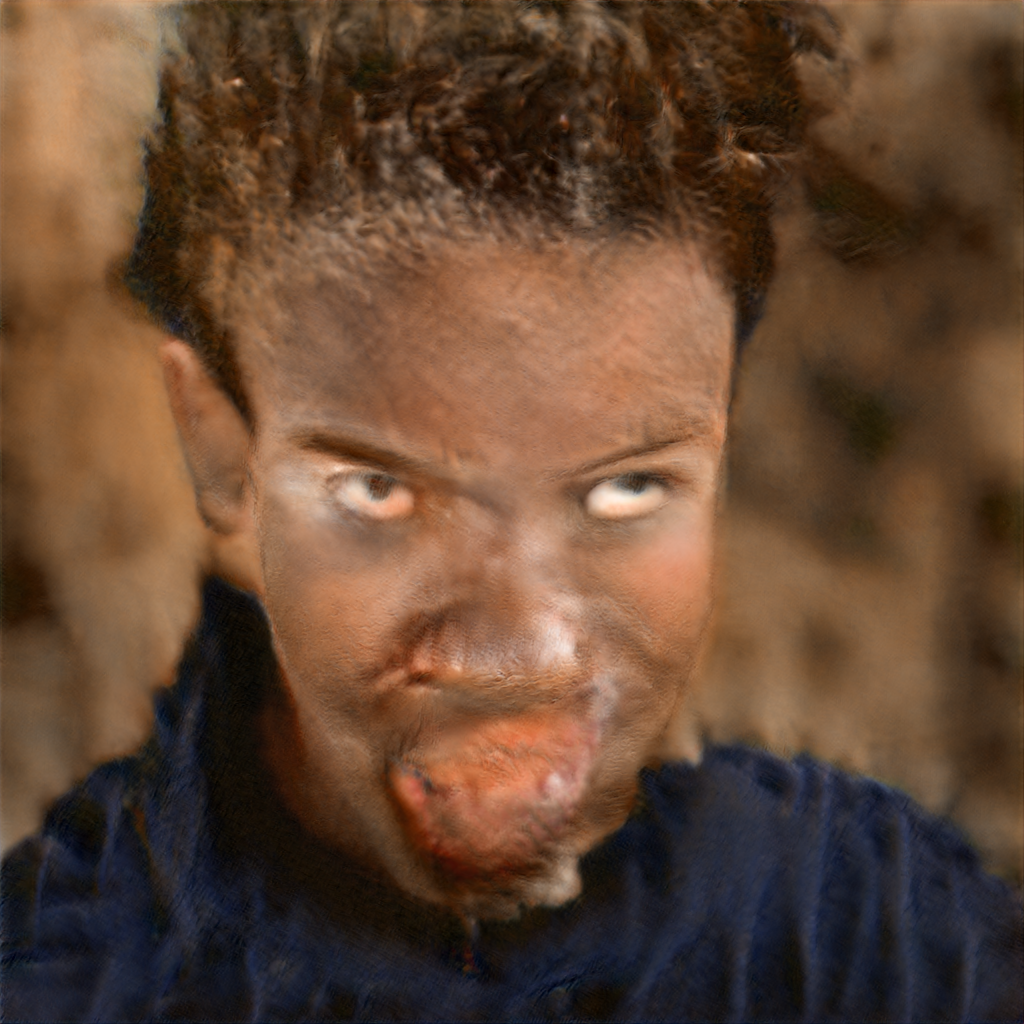}}
  \subfigure[\texttt{Mask}]{\includegraphics[width=1.8cm]{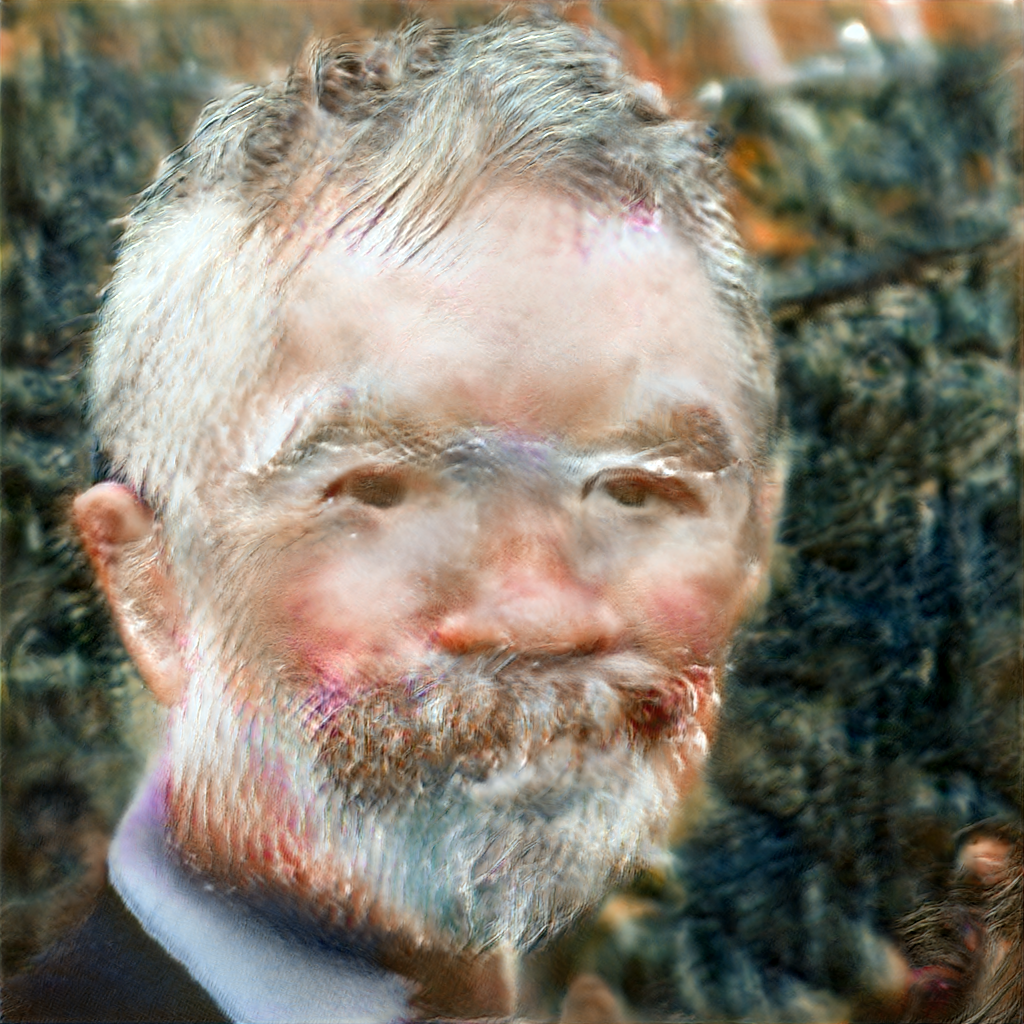}}
  \subfigure[\texttt{Mask}]{\includegraphics[width=1.8cm]{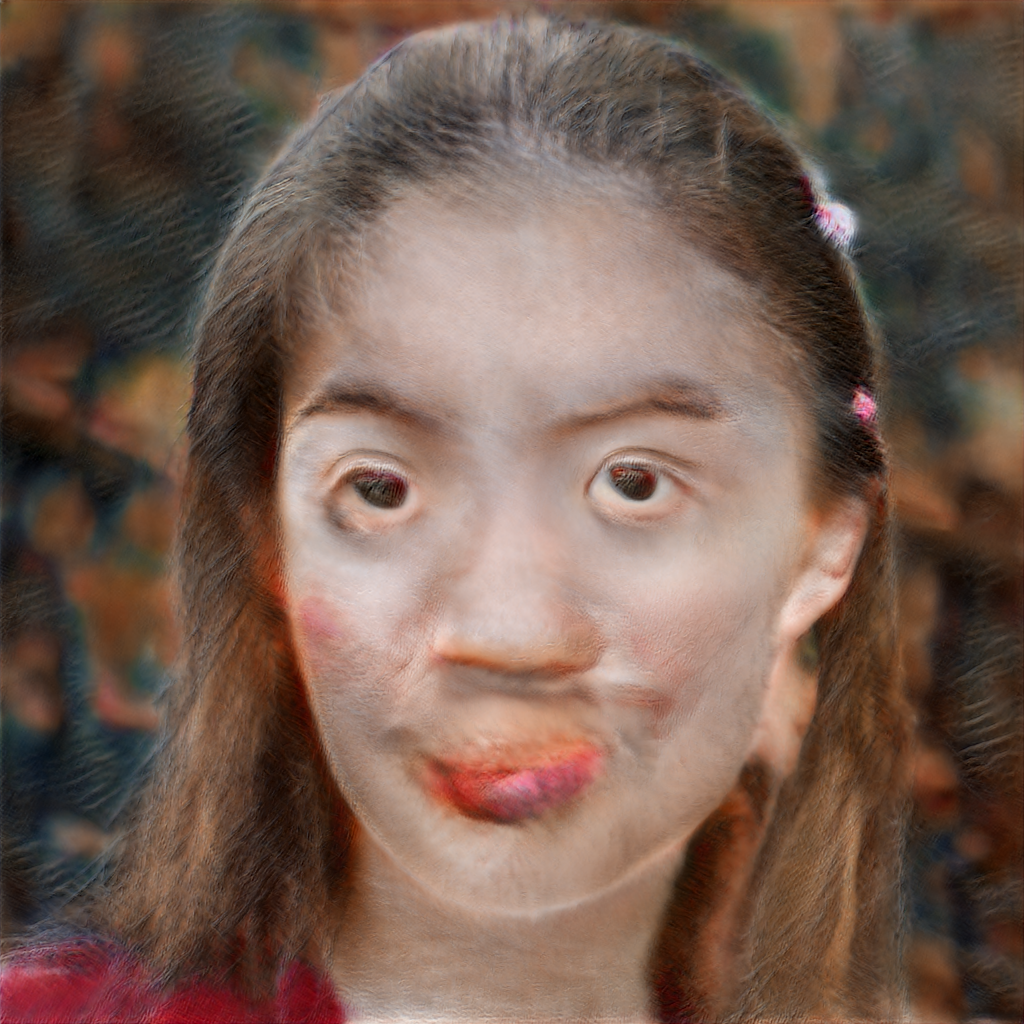}}
  \subfigure[\texttt{Mask}]{\includegraphics[width=1.8cm]{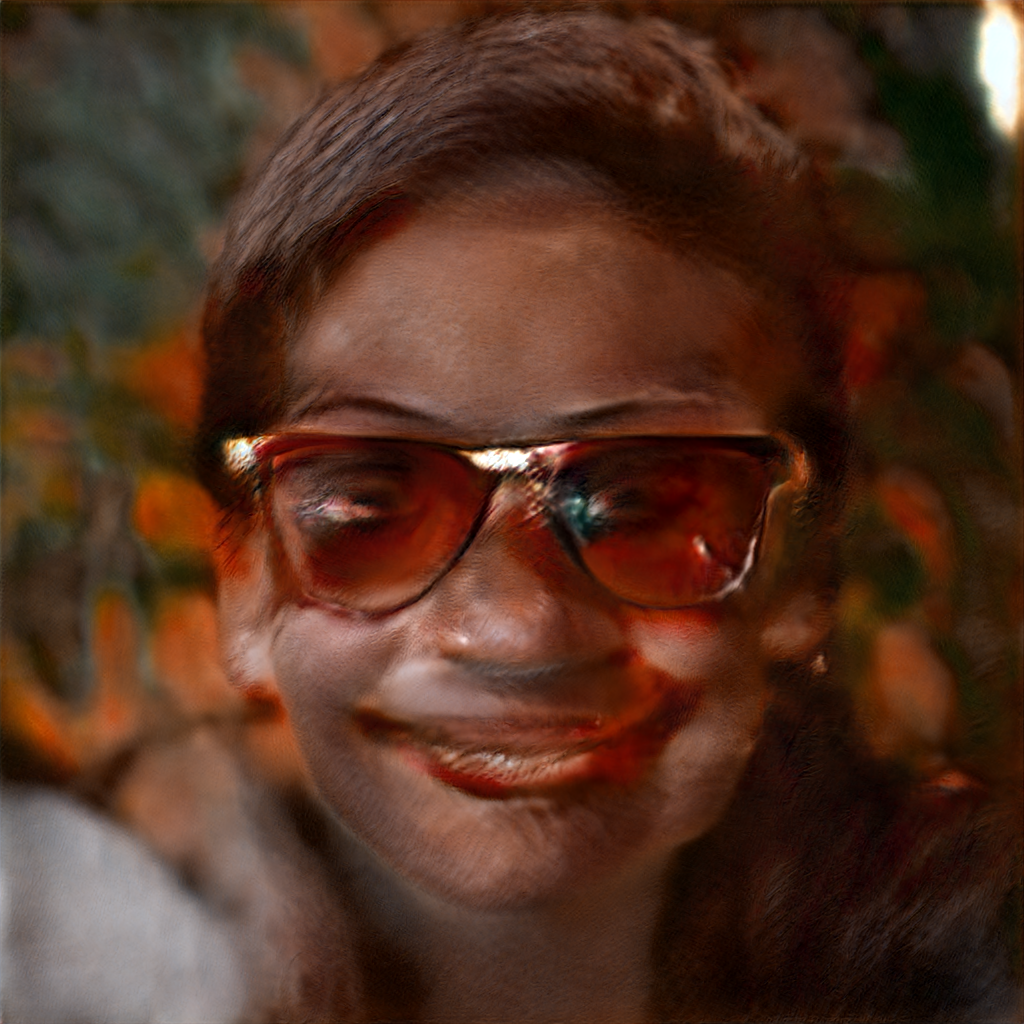}}
  \subfigure[\texttt{Original}]{\includegraphics[width=1.8cm]{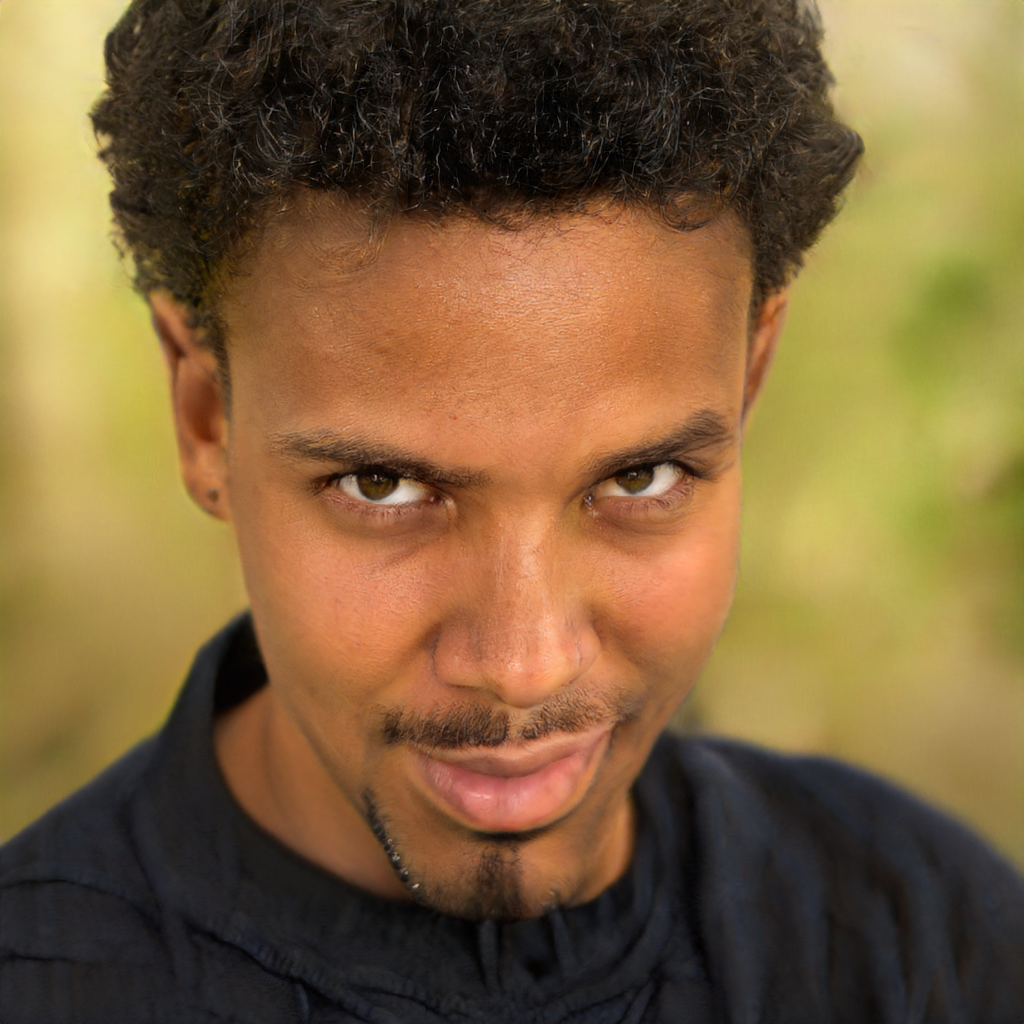}}
  \subfigure[\texttt{Original}]{\includegraphics[width=1.8cm]{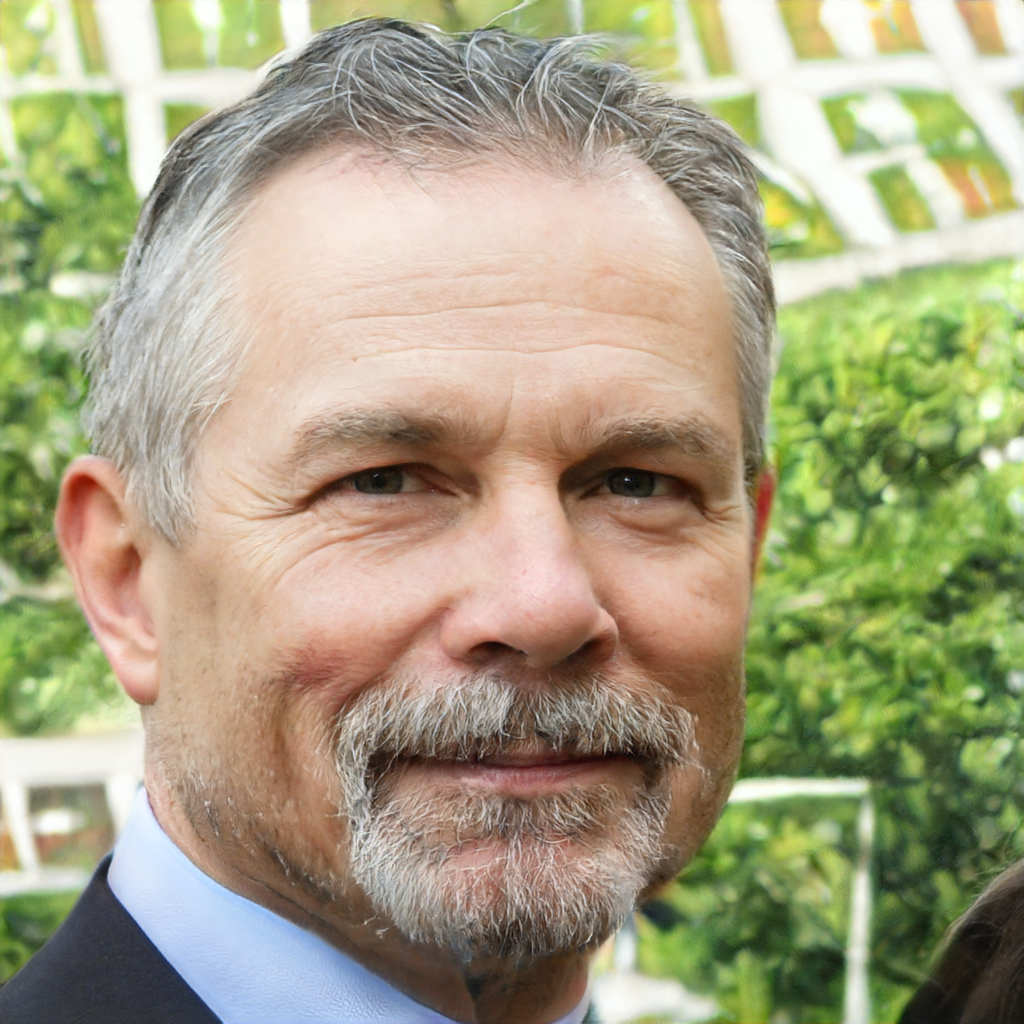}}
  \subfigure[\texttt{Original}]{\includegraphics[width=1.8cm]{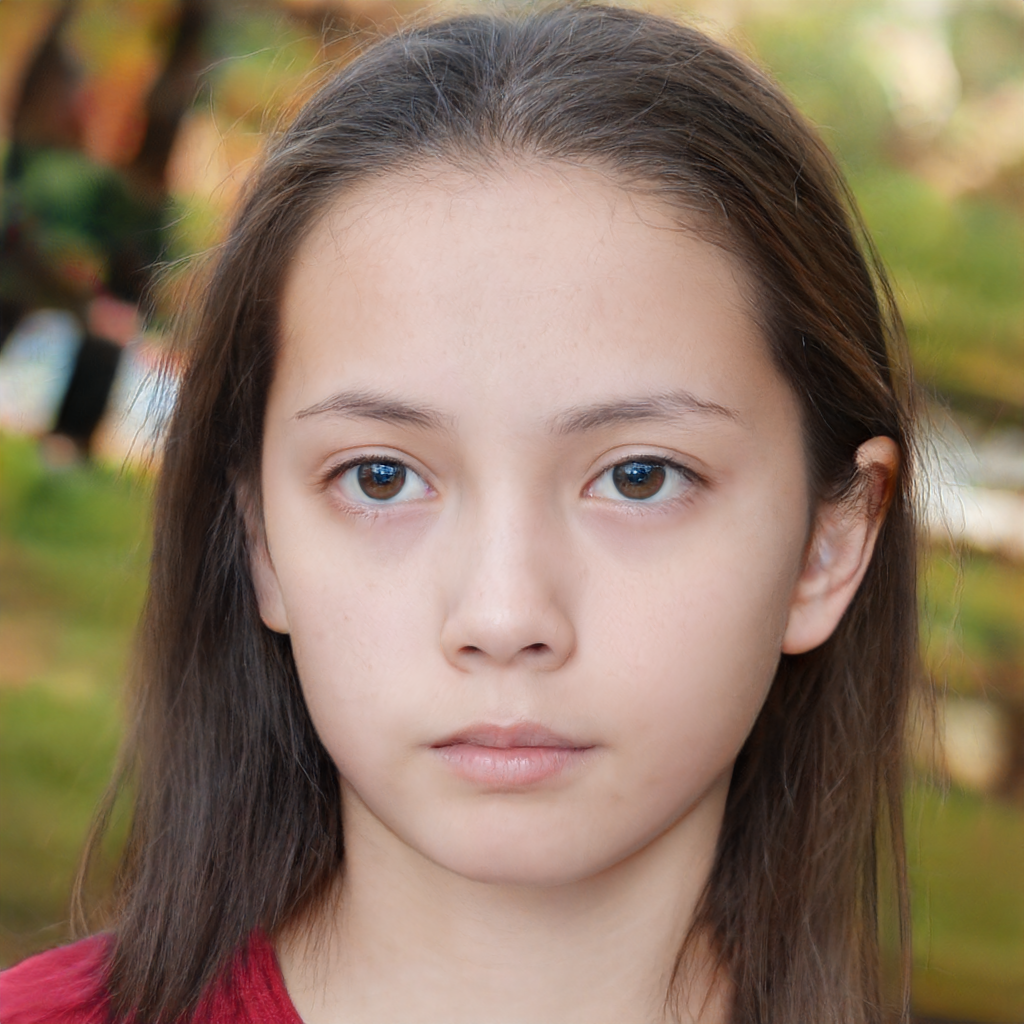}}
  \subfigure[\texttt{Original}]{\includegraphics[width=1.8cm]{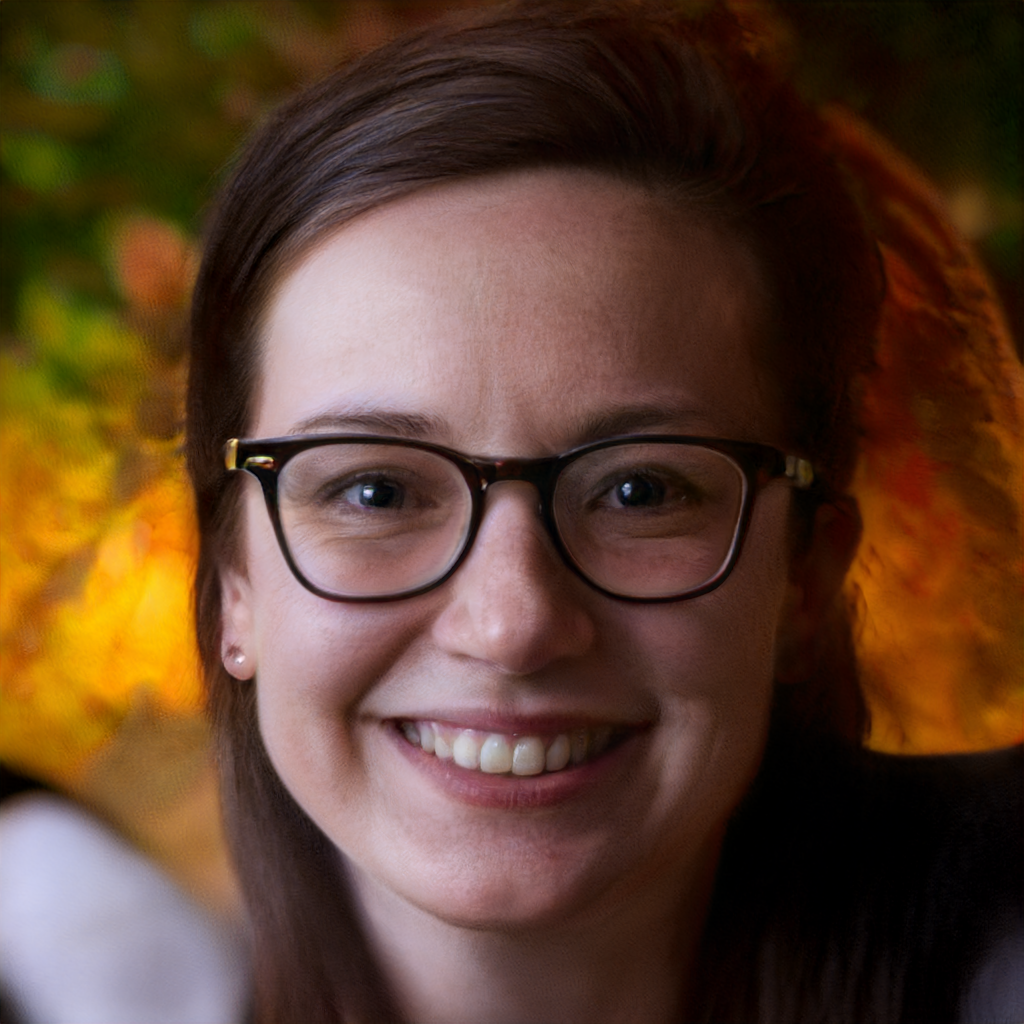}}
  \vspace{-0.1cm}
  \caption{Original images generated via StyleGAN2 and images generated via StyleGAN2 with masked convolutional weights. A column corresponds to the same random seed.}
  \label{Fig: Mask Channels StyleGAN2}
\end{figure}

\begin{figure}
  \centering
  \subfigure[\texttt{}]{\includegraphics[width=3.5cm]{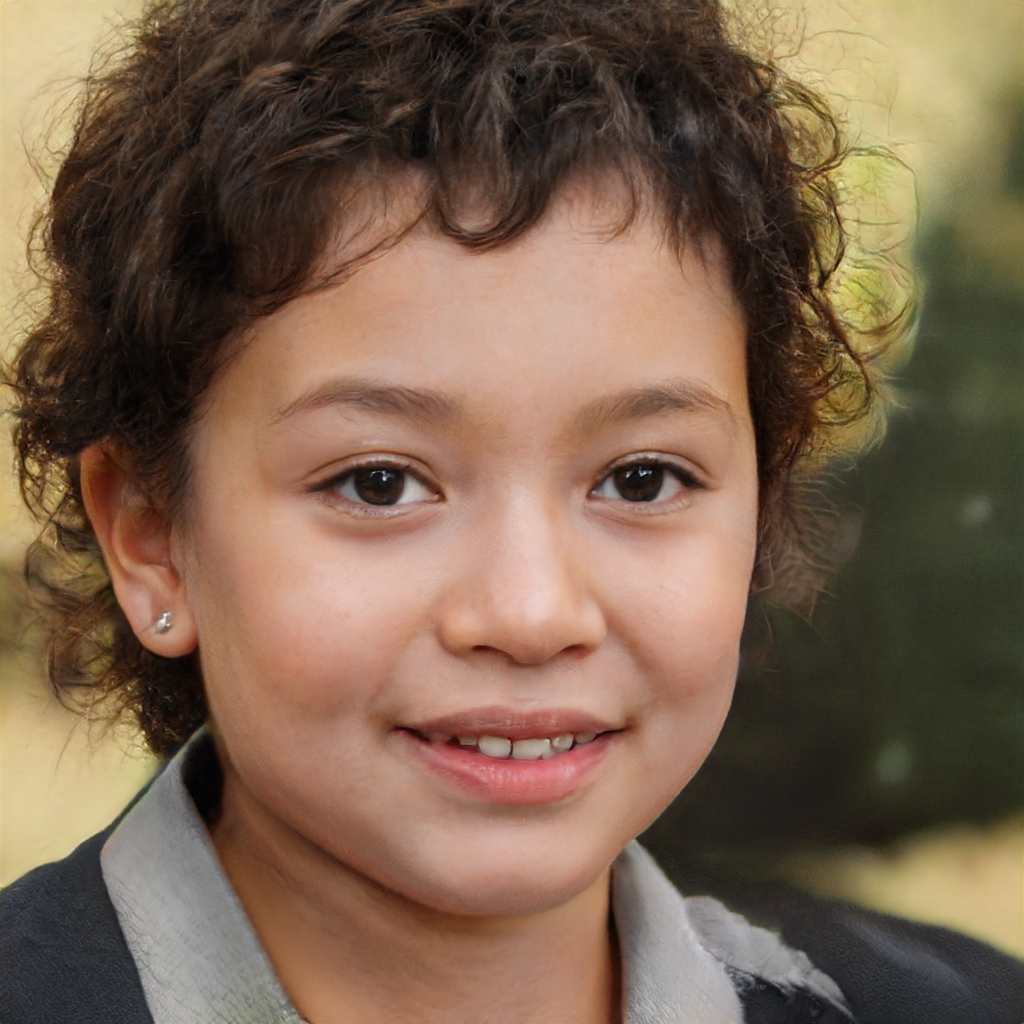}}
  \subfigure[\texttt{}]{\includegraphics[width=3.5cm]{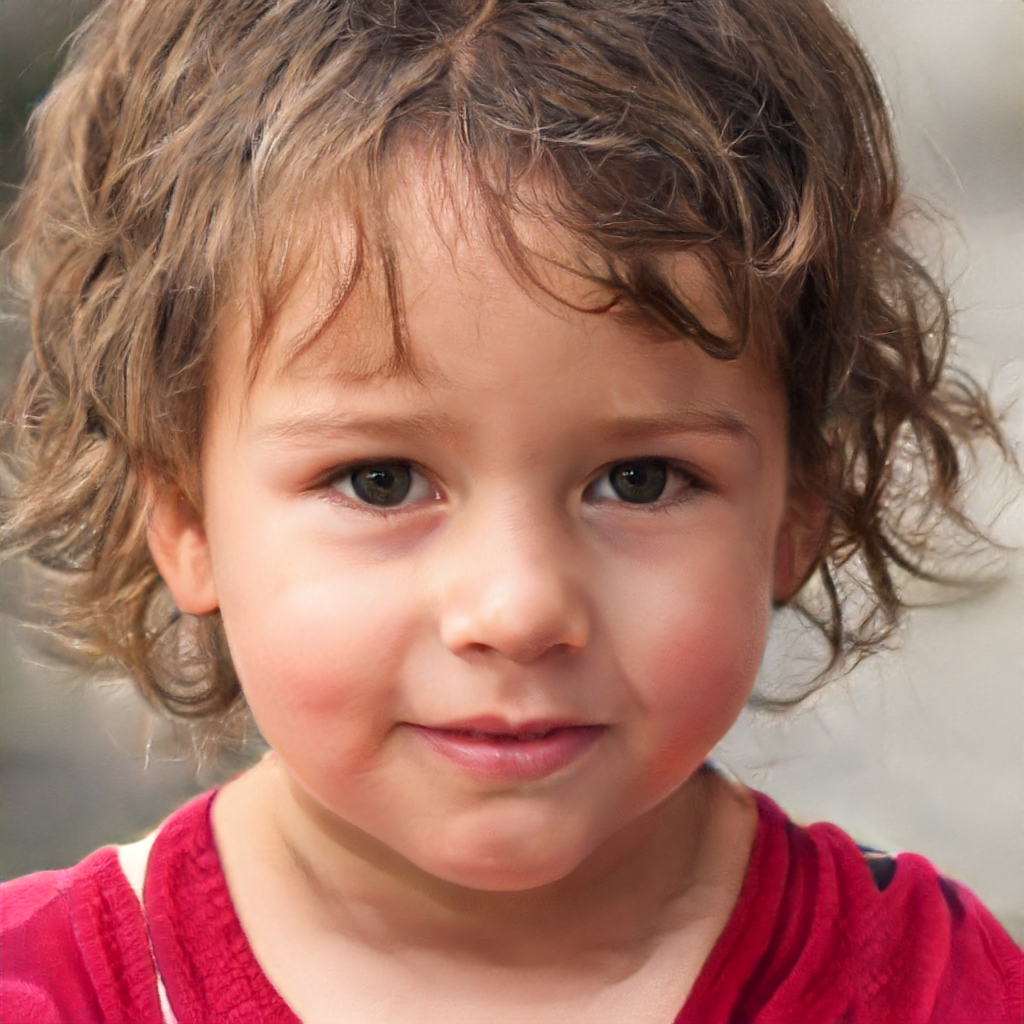}}
  \vspace{-0.3cm}
  \caption{Images generated via StyleGAN2 with identical architectures, which are trained independently, i.e., independent initialization. The two images are synthesized via completely identical latent code $z$.}
  \label{Fig: Different initialized StyleGAN2}
\end{figure}

\item \emph{An individual input lacks the corresponding ground truth in noise-to-image generation task.}\label{difficulty-b} What we want to emphasize is that the ground truth for a vanilla GAN is a distribution of real images, rather than an individual image, which means that a certain latent code sent to the generator may not have the corresponding ground truth. To illustrate that, we send the same latent code $z$ to two independently trained generators of StyleGAN2 and obtain the synthesized images, shown in Fig. \ref{Fig: Different initialized StyleGAN2}. The images are distinct, yet the two generators are with similar performance, i.e., Fréchet Inception Distance (FID). Thus, given that an individual input lacks the corresponding ground truth, aligning the intermediate or output feature maps of the student generator with that of the teacher generator may not be feasible. We need to revisit KD technique in GANs.



\item \emph{The role of a pretrained GAN model in compressing vanilla GANs is unknown.}\label{difficulty-c}
Given the above point, ground truth is absent for an individual input, which means that directly imitating the teacher generator in the feature space or the logits of the teacher discriminator may not be feasible, verified by our experiments in Section \ref{sec:ablation_study}. How to effectively leverage the knowledge of teacher GANs is mysterious. 

\item \emph{Non-convex non-concave Minimax structure of GAN increases the difficulty of the compression.} 
The non-convex non-concave minimax structure causes unstable training of GANs \cite{salimans2016improved, sun2020towards}, imposing an additional challenge on the compression.  
Directly introducing teacher GANs, especially the teacher discriminator, could aggravate the instability of training student GANs since the pretrained teacher discriminator is too strong to provide appropriate supervision for student generator, which has also been demonstrated via experiments in Section \ref{sec:ablation_study}. 
\end{enumerate}

Given the sensitivity of the pretrained generator to the reduction of parameters, we directly train a compressed generator from scratch, rather than tuning from the pretrained generator \cite{liu2021content, lin2021anycost}. We expect that directly training a compressed generator with the specific knowledge distillation mechanism we introduce will alleviate the sensitivity to parameters. Taking difficulty (b) and difficulty (c) into account, what we want to introduce from the teacher GAN is the information about the distribution. rather than an individual synthesized image. Directly obtaining the distribution from the teacher generator is highly non-trivial, due to the high dimension of images. However, the teacher discriminator implicitly contains the information about distribution. Recall the optimal formulation of the discriminator in \cite{goodfellow2014generative},
\begin{align}\label{Eq: Optimal discriminator}
    & D^*(x):=\frac{p_d(x)}{p_d(x)+p_g(x)},
\end{align}
where $p_d$ is the distribution of the real datasets and $p_g$ is the distribution of synthesized images. Supposed that the teacher discriminator reaches or approaches the optimal formulation in Eq. \eqref{Eq: Optimal discriminator} and the teacher generator achieves a superior performance than the student generator (regarding certain divergence metrics, e.g., FID), compared with the student discriminator, the teacher discriminator may contain a better synthesized distribution. A natural question arises, does the teacher discriminator is more suitable for distilling student GANs than the teacher generator? In this work, we answer the above question from empirical perspectives. Specifically, we propose to only exploit the teacher discriminator as the additional source of the supervision signal for the student generator. Moreover, to alleviate the instability caused by the minimax objective function, we propose a two-stage training strategy to stabilize the optimization process.

In a word, our framework only utilizes the teacher discriminator and trains a narrow generator from scratch, with a two-stage training strategy to stabilize the optimization process, denoted as {\bf D}iscriminator {\bf G}uided {\bf L}earning approach and abbreviated as DGL-GAN, aiming at compressing large-scale vanilla GANs and making inference efficient. Our contributions can be summarized as,


\begin{itemize}[leftmargin=*]

\item Motivated by the intuition that the teacher discriminator may contain some useful information about distribution about real images and fake images, we merely exploit the adversarial interaction between the teacher discriminator and student generator to facilitate student generator, which is simple yet effective.


\item We validate DGL-GAN on two representative large-scale vanilla GANs, e.g., StyleGAN2 and BigGAN. DGL-GAN achieves state-of-the-art in compressing both StyleGAN2 and BigGAN. With DGL-GAN, the compressed StyleGAN2 achieves comparable Fréchet Inception Distance (FID) \cite{heusel2017gans} using only $1/3$ parameters on FFHQ, and the compressed BigGAN achieves comparable Inception Score (IS) \cite{salimans2016improved} using only $1/4$ parameters on ImageNet.

\item Not confined to compressed vanilla GANs, DGL-GAN also facilitates the performance of original uncompressed StyleGAN2, achieving a new state-of-the-art performance on FFHQ (FID 2.65), even surpassing StyleGAN3 \cite{karras2021alias}.

\end{itemize}
\section{Related work}\label{related-work-sec}
Model compression has been extensively studied especially for image-classification tasks, see e.g., \cite{han2015deep, han2015learning, hinton2015distilling, courbariaux2015binaryconnect, wu2016quantized}. The typical model compression techniques include weight quantization/sparsification \cite{courbariaux2015binaryconnect, wu2016quantized, han2015deep}, network pruning \cite{hu2016network, li2016pruning}, KD \cite{hinton2015distilling, yim2017gift}, and lightweight neural network architecture/operation design \cite{tan2019efficientnet, liu2018darts, howard2017mobilenets}. In this work, we mainly focus on model compression for vanilla GANs, i.e., noise-to-image task. Most of the existing GAN compression techniques are borrowed from the above-mentioned techniques developed for image-classification. 

\emph{Large-scale vanilla GANs.} Vanilla GANs, targeting at transforming noise into natural images, has been developed and researched thoroughly. However, regarding synthesizing high-resolution images, BigGAN \cite{brock2018large} is the first to synthesize images on ImageNet with resolution 128x128 or 512x512, equipped with class conditional batch normalization, projection discriminator, etc. StyleGAN series \cite{karras2019style, karras2020analyzing, karras2021alias, sauer2022stylegan} are able to synthesize images with even higher resolution, e.g., 1024x1024 FFHQ. Though breakthrough has been achieved via BigGAN and StyleGAN series, the computation cost of them are enormous, due to the wide generator/discriminator (e.g., the bottom of generator of StyleGAN has $512$ channels, the bottom of the generator of BigGAN has $1536$ channels) and the gradually expanded feature map in the spatial dimension. Compressing large-scale vanilla GANs is an urgent but non-trivial task, impeded by the difficulties mentioned in Section \ref{sec:introduction}.

\emph{Lightweight GANs design.}\ 
Liu et al. \cite{liu2021towards} explore a lightweight GAN via a skip-layer channel-wise excitation module and a self-supervised discriminator for compressing StyleGAN2 and achieved remarkable performance. MobileStyleGAN \cite{belousov2021mobilestylegan} utilizes a wavelet-based convolutional neural network and a lightweight depth-wise separable modulated convolution to accelerate StyleGAN2.

\emph{Conditional GANs compression.}\ 
Conditional GANs compression for image-to-image generation task can be divided into the following categories: KD, network pruning, weight quantization, and evolutionary compression. The mainstream of conditional GAN compression \cite{aguinaldo2019compressing,yu2020self,ren2021online,chen2020distilling,li2020gan,jin2021teachers,li2020learning,fu2020autogan,zhang2018self} exploits KD, transferring the knowledge of teacher generator to student generator. Besides KD, several works \cite{chen2021gans,wang2020gan,shen2020cpot} exploit network pruning. In addition, there exist several works \cite{li2021evolutionary,wang2019evolutionary} developing evolutionary compression with inferior results compared with KD-based approaches. 

\newcommand{\tabincell}[2]{\begin{tabular}{@{}#1@{}}#2\end{tabular}}
\begin{table*}[t]
\centering
\small 
  \caption{Comparison between DGL-GAN and previous vanilla GAN compression methods. $G$ and $D$ denote the student generator and the student discriminator, respectively. $\overline{G}$ and $\overline{D}$ denote the teacher generator and the teacher discriminator, respectively.}
  \label{Tab: comparasion_dgl_baselines}
  \begin{tabular}{cccc}
    \toprule
    \multirow{2}*{Name} & \multirow{2}*{Compression} & \multicolumn{2}{c}{Training} \\
    \cline{3-4}
     &  & Target of $G$ & Target of $D$ \\
    \midrule
    Anycost-GAN \cite{lin2021anycost} & Sub-networks & Imitating $\overline{G}$ & \multirow{4}*{Same as common GANs} \\
    CA-GAN \cite{liu2021content} & Pruning & Imitating $\overline{G}$ & \\
    VEM \cite{kang2022information} & Pruning & \tabincell{c}{Maximizing mutual information\\ between $G$ and $\overline{G}$} & \\
    \midrule
    DGL-GAN & Reducing channels & Fooling $D$ and $\overline{D}$ & Same as common GANs \\
    \bottomrule
  \end{tabular}
\end{table*}

\emph{Vanilla GANs compression.}\ 
Recently, some works \cite{aguinaldo2019compressing,yu2020self,lin2021anycost,hou2020slimmable,liu2021content,chen2021gans, li2021revisiting, liu2022sparse} focus on the compression of vanilla GANs, which is more difficult and challenging. Aguinaldo et al. \cite{aguinaldo2019compressing} are the first to apply the KD technique to compress vanilla GANs, which supervises the student generator via the teacher generator for compressing DCGAN \cite{radford2015unsupervised} and WGAN \cite{arjovsky2017wasserstein}. Yu et al. \cite{yu2020self} further propose the objective function to align the loss values of the student generator with that of the teacher generator. Chen et al. \cite{chen2021gans} finds that lottery ticket hypothesis \cite{frankle2018lottery} also holds for GANs and compresses SN-GAN \cite{miyato2018spectral} on CIFAR-10. Slimmable GAN \cite{hou2020slimmable} gradually distills the generator towards the reduction of the channels. CA-GAN \cite{liu2021content} conducts punning over the teacher generator and distillation on the student generator, where the content of interest is fed to the KD scheme. Anycost-GAN \cite{lin2021anycost} compresses StyleGAN2 to edit natural synthesized images in the resource-constrained scenario, equipped with multi-resolution training, adaptive channels, and generator-conditioned discriminator, etc. GCC \cite{li2021revisiting} explores the student discriminator with flexible masks of convolutional weights during the compression of GANs. Kang et al. \cite{kang2022information} proposes to maximize the variational lower bound of the mutual information between the student generator and the teacher generator, with the aid of an additional Energy Based Model (EBM). The tendency of the research on the compression of vanilla GANs is towards compressing large-scale vanilla GANs on high-resolution datasets, i.e., from SN-GAN on CIFAR-10 to StyleGAN2 on FFHQ or BigGAN on ImageNet, which is non-trivial to be compressed. We summarize the differences between our method and previous methods in Tab. \ref{Tab: comparasion_dgl_baselines}.

In this work, we focus on vanilla GAN compression on two of the most representative and challenging noise-to-image generative models: StyleGAN2 \cite{karras2020analyzing} and BigGAN \cite{brock2018large}. Our approach is simple yet effective and falls into a teacher-student learning  paradigm. Our approach is different from previous vanilla GANs compression methods in three aspects. 
{\bf First}, instead of directly applying KD to compressing GANs \cite{aguinaldo2019compressing, yu2020self}, DGL-GAN revisits the role of teacher GAN for optimizing student GAN. Motivated by that the teacher discriminator may contain more meaningful information and the intrinsic property of noise-to-image tasks, DGL-GAN transfers the knowledge from the teacher discriminator to student GAN via the formulation of the adversarial objective function of GANs.
{\bf Second}, compared with CA-GAN \cite{liu2021content}, Anycost-GAN \cite{lin2021anycost}, and VEM \cite{kang2022information}, DGL-GAN directly trains a narrow GAN from scratch, instead of tuning from the original pretrained StyleGAN2 model, providing some insights about training a narrow GAN. 
{\bf Third}, compared with GAN-LTH \cite{chen2021gans}, DGL-GAN directly reduces the number of channels in GAN and trains the narrow GAN, rather than obtaining sparse convolutional weights through masks, which is more effective in training and inference.

\section{Vanilla GANs Compression: DGL-GAN}\label{method-sec}
In this section, we propose our novel discriminator-guided learning for vanilla GAN compression, dubbed DGL-GAN. This section is organized as follows. In Section \ref{sec: Reduce width}, we illustrate the method to compress GANs. In section \ref{sec: Motivation}, we illuminate our motivation to propose DGL-GAN. In section \ref{sec:vanillagan}, we revisit the vanilla GAN formulations and the proposed DGL-GAN; 
in section \ref{sec:two-stage-training}, we propose a two-stage training strategy technique to stabilize the training process of DGL-GAN.

\subsection{Reduce the width}\label{sec: Reduce width}
Taking a deep look at the network architectures of BigGAN and StyleGAN2, we can find that the large width and depth cause the increasing parameter size of the networks. Regarding the depth of the network, given the correlation between the resolution of the datasets (e.g., 128x128 for ImageNet or 1024x1024 for FFHQ) and the number of blocks in the generator\footnote{Usually, the number of blocks in the generator can be calculated via $\log_{2}(\frac{{\rm resolution}}{4})$, where $4$ is the width and height of the latent code $z$ received by the generator.}, the depth of the network is difficult to be reduced. Regarding the width of the network, it can be reduced without affecting the generation of images. Thus, we tend to compress the network of large-scale vanilla GANs by directly reducing the width. 

Distinct from sparse methods \cite{chen2021gans} and pruning methods \cite{liu2021content}, which utilizes a sparse mask $m$ to disable some elements in the convolutional weights $w$, DGL-GAN directly explores the dense narrow network. In DGL-GAN, supposed that the convolution weight $w_o$ in original GANs with shape $[{\rm C_{out}}, {\rm C_{in}}, {\rm K_h}, {\rm K_w}]$, the shape of the corresponding weight $w_c$ in compressed GANs is $[{\rm C_{out}}\times($Ch-Mul$), {\rm C_{in}}\times($Ch-Mul$), {\rm K_h}, {\rm K_w}]$, where Ch-Mul is denoted as the channel multiplier. Thus, reducing the width compresses both the convolution weights $w_c$ and the intermediate features, while the sparse model does not compress the intermediate features during training and inference, thus DGL-GAN is GPU-friendly and more efficient during training and inference than sparse methods \cite{chen2021gans}. 

\subsection{Motivation}\label{sec: Motivation}
We start with vanilla GAN \cite{goodfellow2014generative}, which is formulated as the following minimax optimization problem:
\begin{align}\label{original function}
\small & \min_{G}\max_{D} {\rm Adv}(G,D)\nonumber \\
& \min_{G}\max_{D} \mathbb{E}_{x\sim p_d}[f(D(x))]\!+\!\mathbb{E}_{z\sim p_z}[g(D(G(z)))]\nonumber \\
& \min_{G}\max_{D} \mathbb{E}_{x\sim p_d}[f(\mathcal{A}(h(x)))]\!+\!\mathbb{E}_{z\sim p_z}[g(\mathcal{A}(h(G(z))))],
\end{align}
where ${\rm Adv}(G,D)$ can be non-saturating loss in \cite{goodfellow2014generative} or hinge loss in \cite{miyato2018spectral} by taking various loss functions $f$ and $g$. $p_d$ is the distribution of real data, and $p_z$ is the distribution of the latent code, usually Gaussian distribution. $D(x)$ can be written as $\mathcal{A}(h(x))$, where $h(x)$ and $h(G(z))$ calculate the logits for the real data and fake data respectively, and $\mathcal{A}$ is an activation function to obtain the probability of the real data and the fake data respectively, usually sigmoid function.

Given the correlation between the capacity of the neural network and the number of parameters \cite{hornik1991approximation, allen2019convergence}, with the compression of the neural network, the performance drop is inevitable, shown in Fig. \ref{Fig: Performance_curve}. How to alleviate the performance drop of the compressed network is the key problem we want to address. In the scope of our research, we want to explore how to effectively introduce the supervision signal from teacher GAN to boost the performance of student GAN.

It is non-trivial to implement the teacher-student learning of vanilla GANs due to the difficulties mentioned in Section \ref{sec:introduction}. As illustrated in Section \ref{sec:introduction}, the teacher discriminator implicitly contains the distributions of real data and images synthesized via the teacher generator. Then how to utilize the implicitly contained distributions in the teacher discriminator? Inspired from the original objective function Eq. \eqref{original function} of GANs and the adversarial training\cite{goodfellow2014explaining, bai2021recent, bai2022improving}, we propose to replace the student discriminator in ${\rm Adv}(G,D)$ with the teacher discriminator, written as 
\begin{align}\label{Eq: DGL-term}
    & \min_{G} {\rm Adv}(G,\overline{D}):=\min_{G} \mathbb{E}_{x\sim p_z}[g(\overline{D}(G(z)))]\nonumber
\\&=\min_{G} \mathbb{E}_{x\sim p_z}[g(\mathcal{A}(\overline{h}(G(z))))],
\end{align}
where $\overline{D}$ denotes the teacher discriminator with the final activation function and $\overline{h}$ denotes the teacher discriminator before the final activation function. 


\begin{figure}[t]
  \centering
  \includegraphics[width=0.4\textwidth]{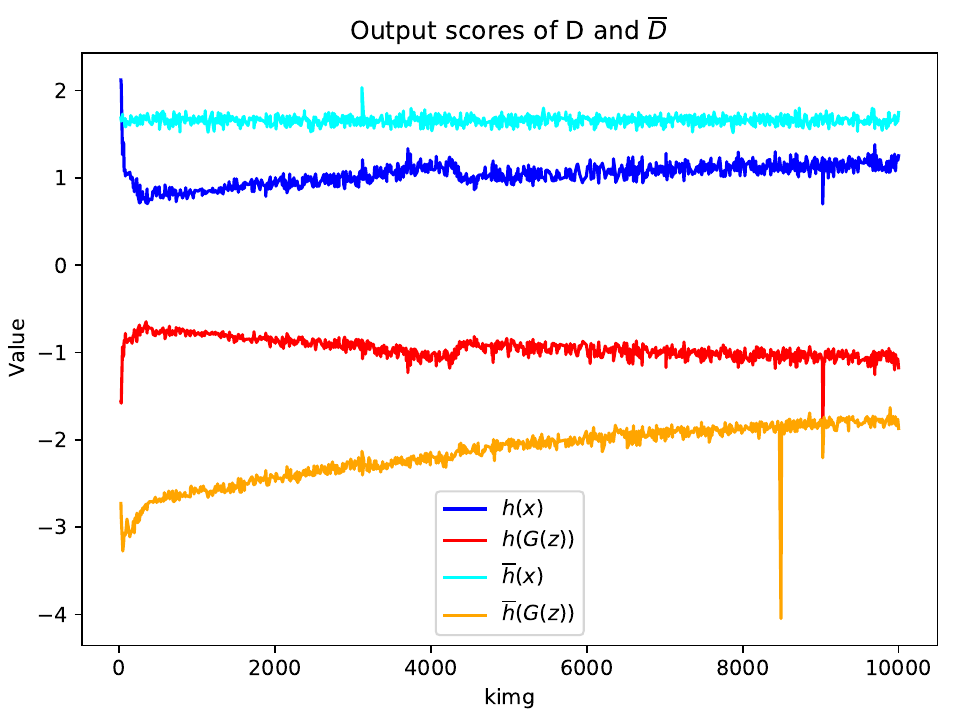}
  \vspace{-0.3cm}
  \caption{The curves of output scores of the student discriminator $D$ with channel multiplier $1/2$ and the pretrained teacher discriminator $\overline{D}$.}
  \label{Fig: h(x) and hbar(x)}
\end{figure}

To analyze the above question, we firstly observe the behavior of student GANs with the channel multiplier $1/2$ and teacher GANs during training. As illustrated in \cite{karras2020training}, the output scores $h(x)$ and $h(G(z))$ of the discriminator $D$ is indicative in inspecting the training status of GANs. We optimize the objective function Eq. \eqref{original function}, and send generated images $G(z)$ and real images $x$ to the teacher discriminator $\overline{D}$, which is the official released model\footnote{We obtain the released model of StyleGAN2 in \href{https://github.com/NVlabs/stylegan2}{https://github.com/NVlabs/stylegan2}} of StyleGAN2 \cite{karras2020analyzing}. The teacher discriminator $\overline{D}$ is not updated during the training process. We plot the curves of $h(x)$ and $h(G(z))$ of $D$ and $\overline{h}(x)$ and $\overline{h}(G(z))$ of $\overline{D}$, presented in Fig. \ref{Fig: h(x) and hbar(x)}. It can be seen that $\overline{h}(x)>h(x)$ and $\overline{h}(G(z))<h(G(z))$, indicating that the teacher discriminator $\overline{D}$ is more confident than the student discriminator $D$. Thus, the teacher discriminator $\overline{D}$ may provide more indicative supervision signal than the student discriminator $D$ in ${\rm Adv}(G,\overline{D})$.

\begin{figure*}[ht]
  \centering
  \begin{minipage}{0.23\textwidth}
    \centering
    \includegraphics[width=0.99\textwidth]{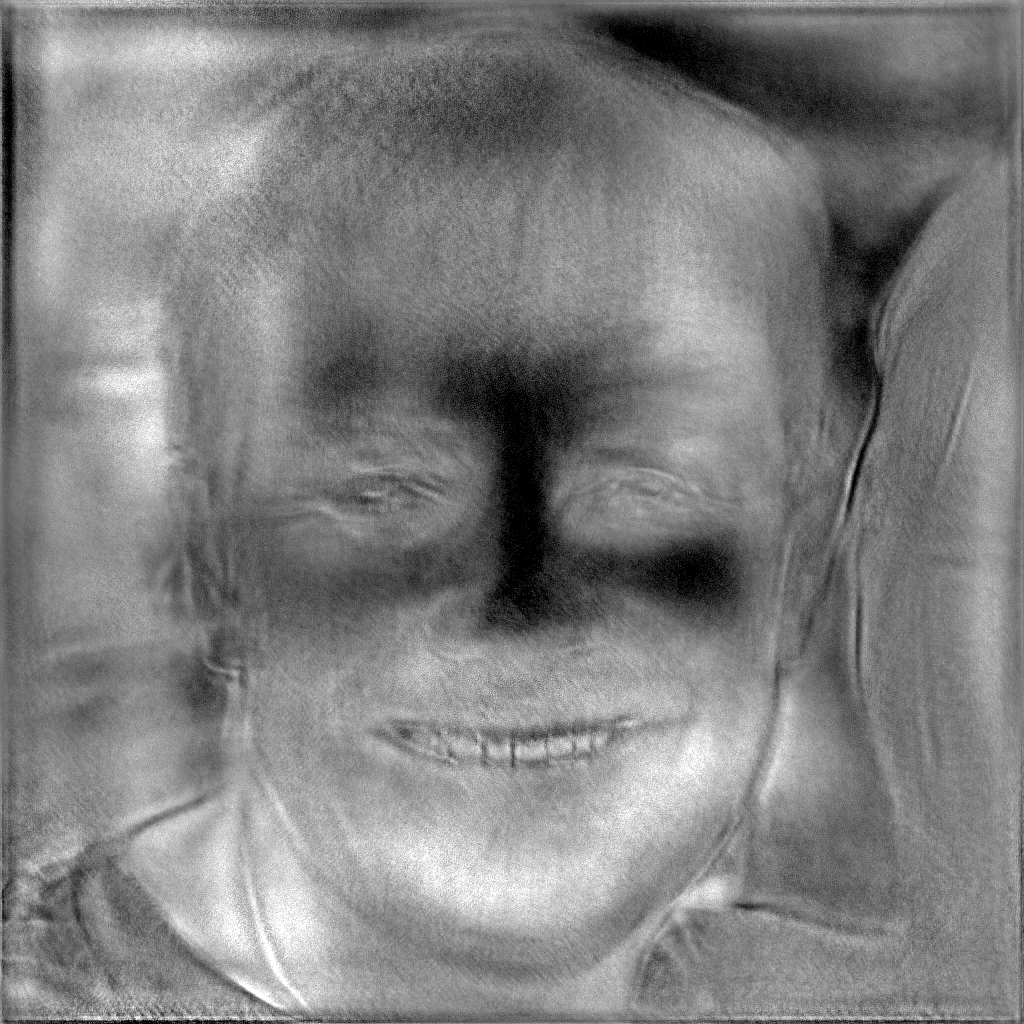}
    \vspace{-0.1cm}
    \centerline{R: $\nabla_{G(z)}{h(G(z))}$}
    \vspace{0.1cm}
  \end{minipage}
  \begin{minipage}{0.23\textwidth}
    \centering
    \includegraphics[width=0.99\textwidth]{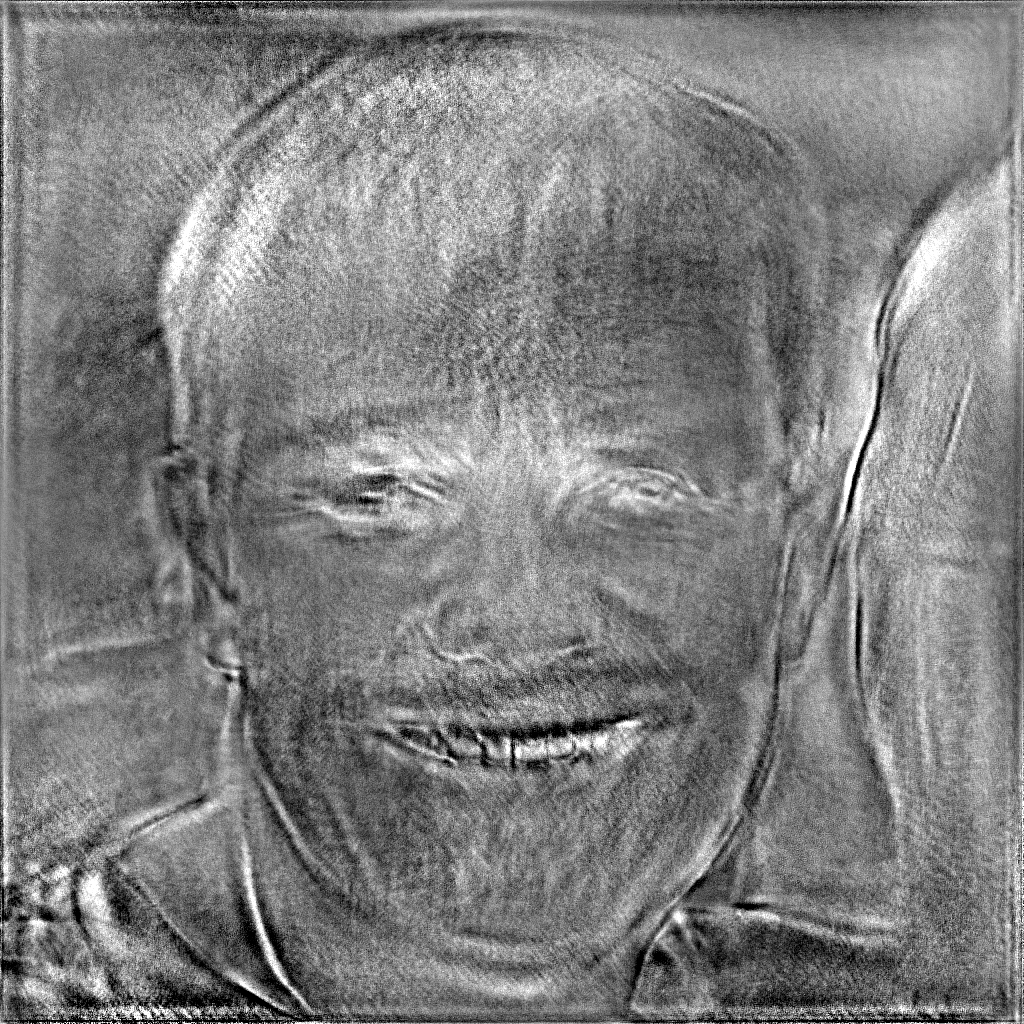}
    \vspace{-0.1cm}
    \centerline{G: $\nabla_{G(z)}{h(G(z))}$}
    \vspace{0.1cm}
  \end{minipage}
  \begin{minipage}{0.23\textwidth}
    \centering
    \includegraphics[width=0.99\textwidth]{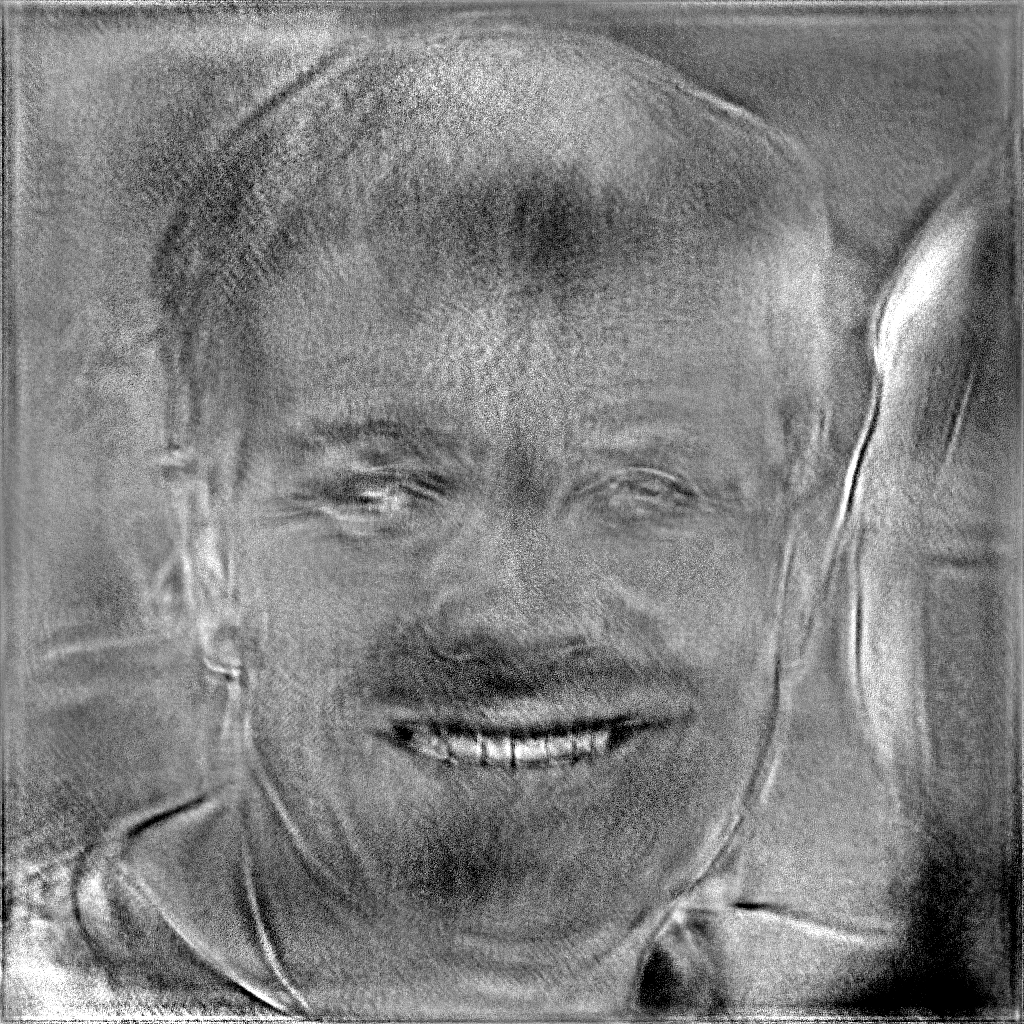}
    \vspace{-0.1cm}
    \centerline{B: $\nabla_{G(z)}{h(G(z))}$}
    \vspace{0.1cm}
  \end{minipage}
  \begin{minipage}{0.23\textwidth}
    \centering
    \includegraphics[width=0.99\textwidth]{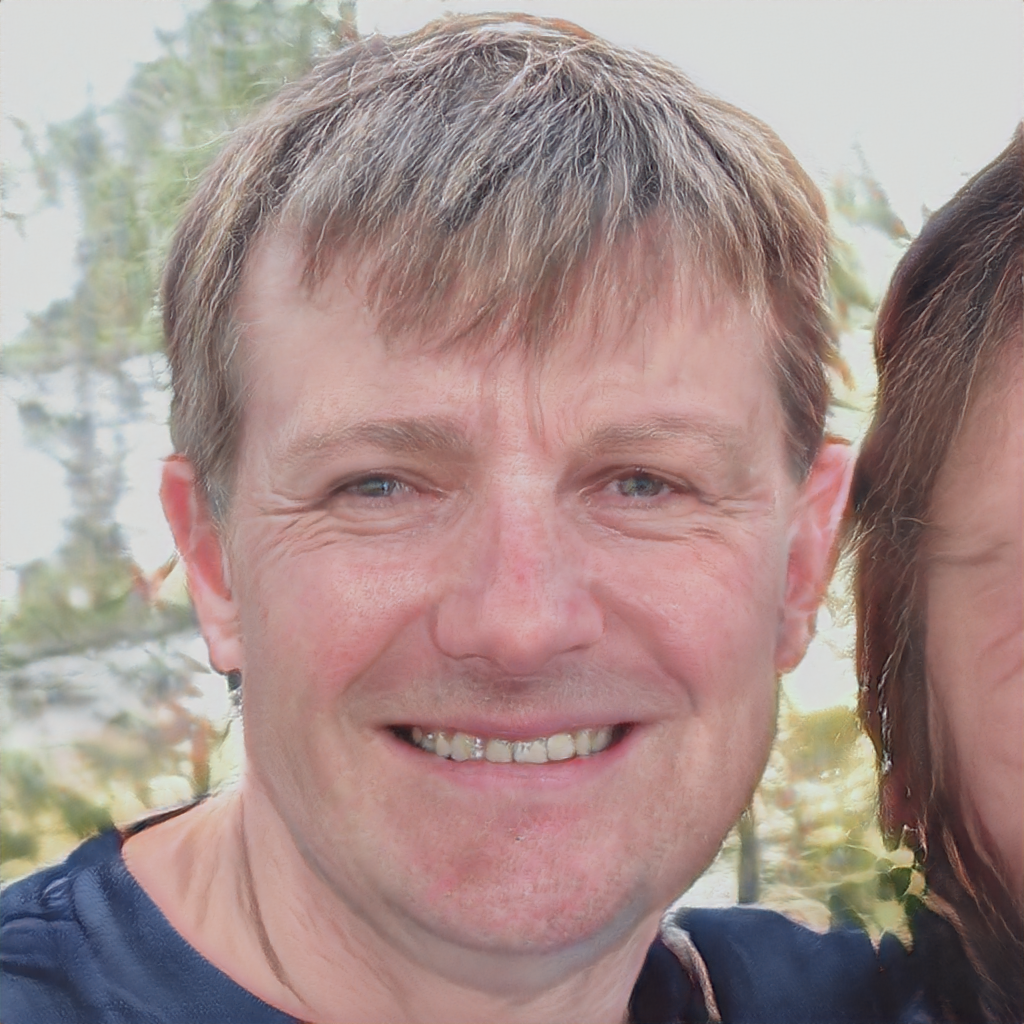}
    \vspace{-0.1cm}
    \centerline{$G(z)$}
    \vspace{0.1cm}
  \end{minipage}
  \begin{minipage}{0.23\textwidth}
    \centering
    \includegraphics[width=0.99\textwidth]{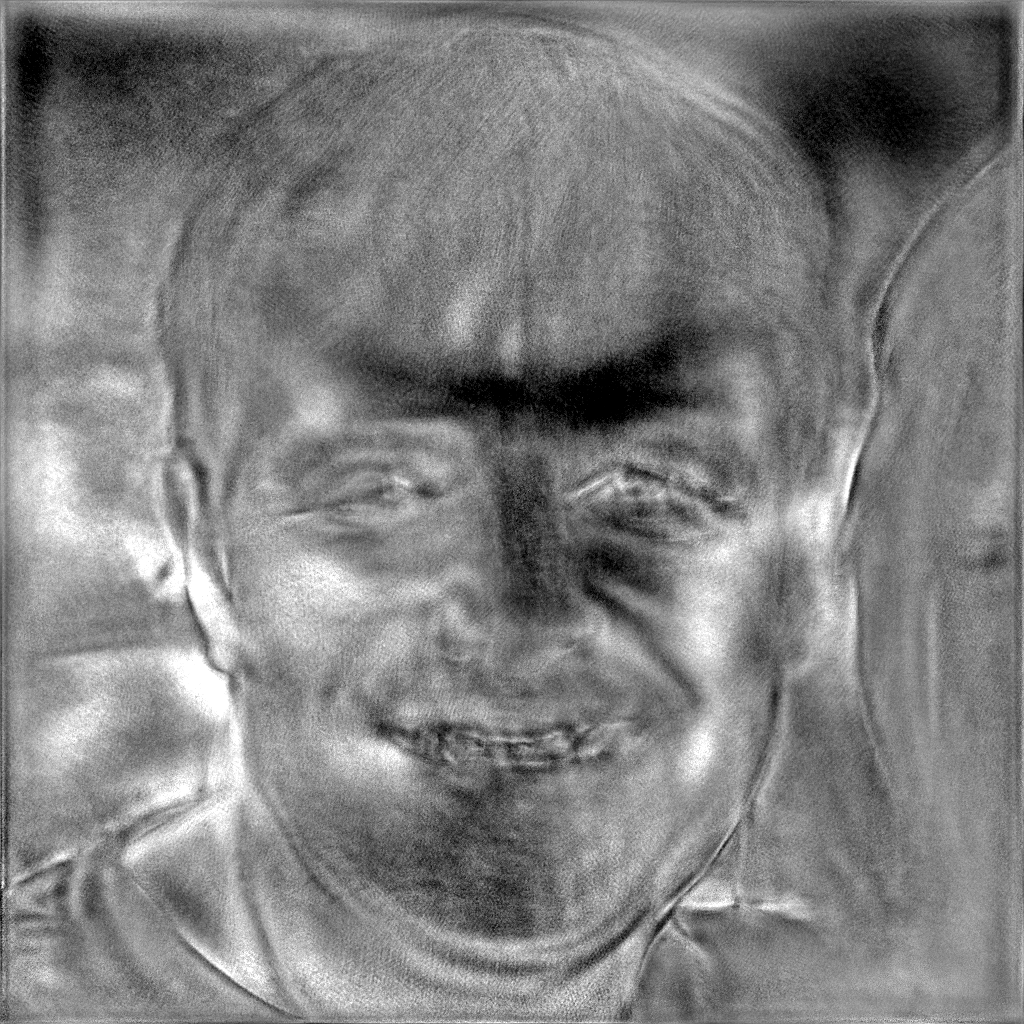}
    \vspace{-0.1cm}
    \centerline{R: $\nabla_{G(z)}{\overline{h}(G(z))}$}
    \vspace{0.1cm}
  \end{minipage}
  \begin{minipage}{0.23\textwidth}
    \centering
    \includegraphics[width=0.99\textwidth]{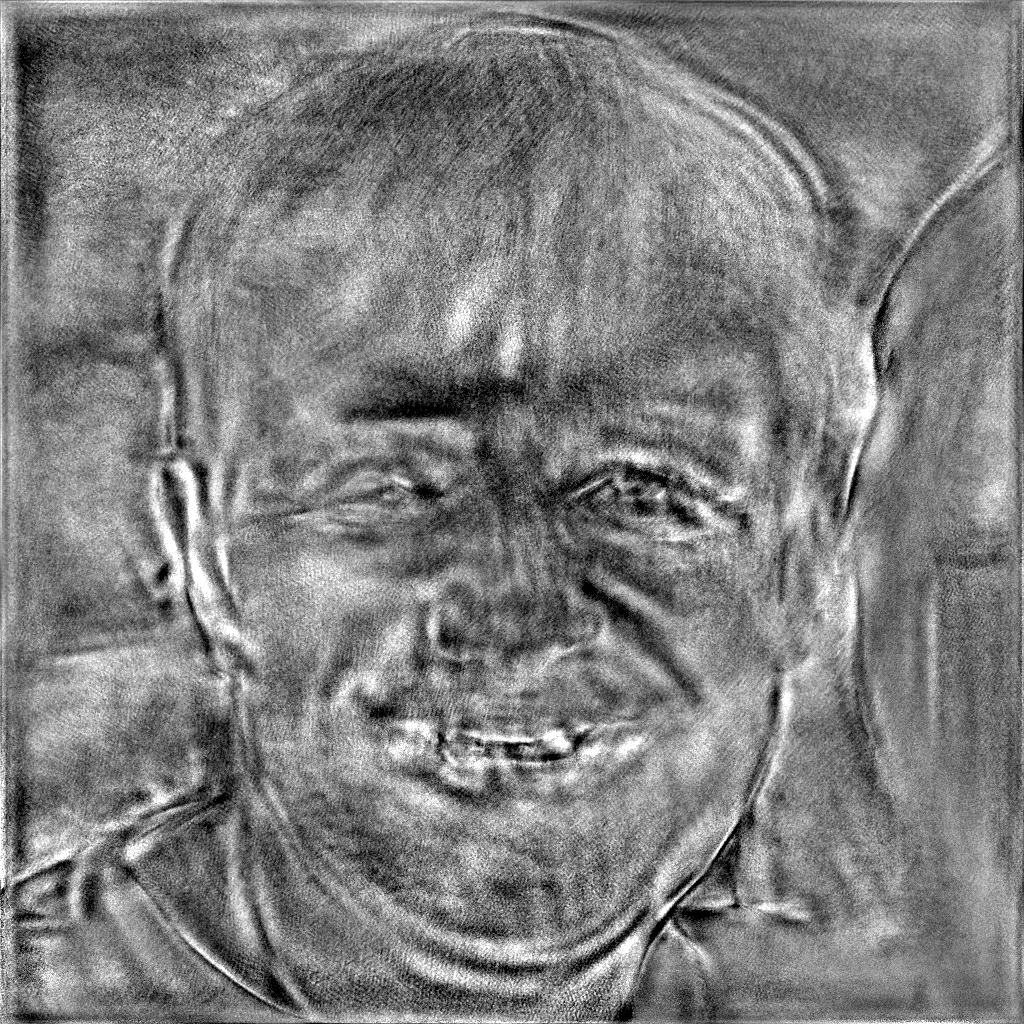}
    \vspace{-0.1cm}
    \centerline{G: $\nabla_{G(z)}{\overline{h}(G(z))}$}
    \vspace{0.1cm}
  \end{minipage}
  \begin{minipage}{0.23\textwidth}
    \centering
    \includegraphics[width=0.99\textwidth]{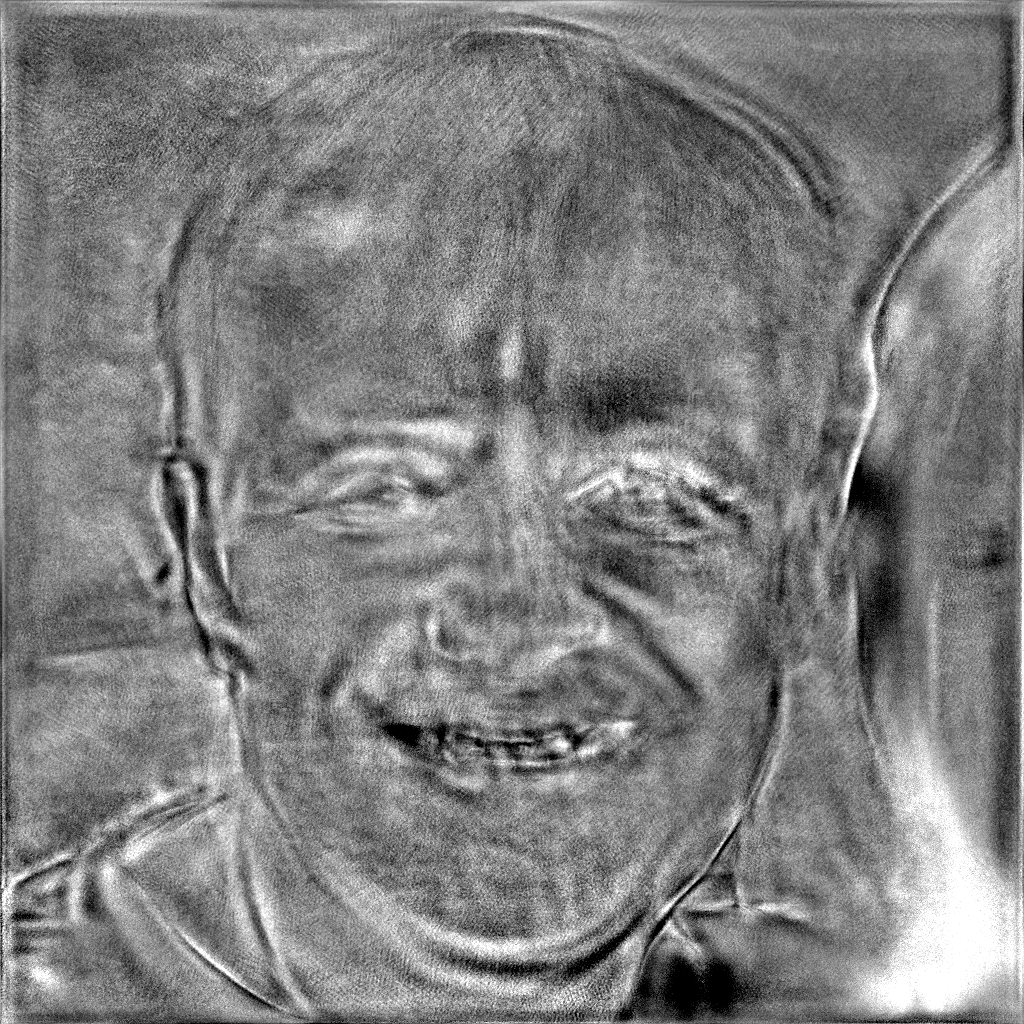}
    \vspace{-0.1cm}
    \centerline{B: $\nabla_{G(z)}{\overline{h}(G(z))}$}
    \vspace{0.1cm}
  \end{minipage}
  \begin{minipage}{0.23\textwidth}
    \centering
    \includegraphics[width=0.99\textwidth]{figures/1-4ch_gradient/fake_1.png}
    \vspace{-0.1cm}
    \centerline{$G(z)$}
    \vspace{0.1cm}
  \end{minipage}
  \caption{The gradients $\nabla_{G(z)}{h(G(z))}$,  $\nabla_{G(z)}{\overline{h}(G(z))}$, and the fake image $G(z)$, better to be viewed via zooming in. We separately visualize the gradient of three channels (i.e., R, G, and B) in $G(z)$, where the gradient values out of $\pm3$ standard deviations are omitted. Note that the gradients $\nabla_{G(z)}{h(G(z))}$ and $\nabla_{G(z)}{\overline{h}(G(z))}$ are calculated under the same input fake image $G(z)$, shown in (d) and (h).}
  \label{Fig: The gradients of D and Dbar}
\end{figure*}

More specifically, to illuminate the impact of $D$ and $\overline{D}$ on $G$, we visualize the gradients $\nabla_{G(z)}{h(G(z))}$ and $\nabla_{G(z)}{\overline{h}(G(z))}$ via SmoothGrad \cite{smilkov2017smoothgrad} in objective functions ${\rm Adv}(G,D)$ and ${\rm Adv}(G,\overline{D})$ during the optimization process of ${\rm Adv}(G,D)$, respectively, shown in Fig. \ref{Fig: The gradients of D and Dbar}. Visually, the gradient $\nabla_{G(z)}{\overline{h}(G(z))}$ of the teacher discriminator $\overline{D}$ is clearer and more indicative than $\nabla_{G(z)}{h(G(z))}$, especially in some regions, e.g., the eyes and the mouth, manifesting that the gradient of the teacher discriminator $\overline{D}$ is somewhat meaningful to the student generator $G$ and perhaps can be the complementary supervision signal to strengthen $G$.

\subsection{The formulation of DGL-GAN}\label{sec:vanillagan}
On the basis of the above analysis, we propose to transfer the knowledge of the teacher discriminator $\overline{D}$ to the student generator $G$ via the adversarial objective function ${\rm Adv}(G,\overline{D})$ in Eq. \eqref{Eq: DGL-term}. Combined with the formulation of vanilla GANs, the formulation of DGL-GAN can be written as 
\begin{align}\label{full_objective}
\small \min_{G} \max_{D} {\rm V}(G,D) =  {\rm Adv}(G,D)+\lambda*{\rm Adv}(G,\overline{D}), 
\end{align}
where $\overline{D}$ is the pretrained teacher discriminator, and $(G,D)$ are the student generator and the student discriminator, which take similar architectures as the teacher generator and the teacher discriminator except with a much smaller number of channels. $\lambda$ is a hyper-parameter to balance two terms in the objective function of DGL-GAN. The term ${\rm Adv}(G,\overline{D})$ can be viewed as a regularization term, aiming at providing the complementary supervision signal that the student discriminator $D$ cannot offer at some cases shown in Fig. \ref{Fig: The gradients of D and Dbar}. Furthermore, the two terms in Eq.~\eqref{full_objective} have a consistent magnitude, which makes the hyperparameter $\lambda$ easily to be tuned.

Notably, the teacher generator $\overline{G}$ is not exploited in DGL-GAN, distinct from previous GAN compression methods \cite{liu2021content, yu2020self}. We want to illustrate the discarding of $\overline{G}$ from two perspectives. First, regarding enforcing the student generator $G$ to imitate the teacher generator $\overline{G}$ in the feature space or the final output images, as demonstrated in Section \ref{sec:introduction}, there is no definition of the ground-truth for an individual input in vanilla GANs. We want to argue that simply aligning the outputs or the intermediate features of $G$ and $\overline{G}$ is not the best way to distill student GANs, also verified by our results in Section \ref{sec:ablation_study}. Second, similar to the term ${\rm Adv}(G,\overline{D})$, can student GANs also be distilled from the term ${\rm Adv}(\overline{G},D)$? In the following section \ref{sec:ablation_study}, we will demonstrate the negative impact of introducing ${\rm Adv}(\overline{G},D)$ on student GANs from the perspective of empirical study.

\subsection{Two-Stage Training Strategy for DGL-GAN}\label{sec:two-stage-training}

DGL-GAN in Eq.~\eqref{full_objective} also takes the minimax structure, in which the student generator $G$ and the student discriminator $D$ are learned adversarially. 
In DGL-GAN, student generator $G$ is simultaneously supervised by its discriminator $D$ and pretrained teacher discriminator $\overline{D}$. The teacher discriminator $\overline{D}$ is trained sufficiently to discriminate two distributions (i.e., real data distribution $p_{d}$ and fake distribution $\overline{p_g}$ produced by the teacher generator $\overline{G}$) with a relatively low divergence (e.g., Jensen-Shannon divergence). However, the fake distribution $p_g$ of the student generator $G$ is far from the data distribution $p_{d}$, especially at the beginning of training DGL-GAN. 
Thus, there may exist a large gap between the student discriminator $D$ and pretrained teacher discriminator $\overline{D}$ at the beginning of training, which leads to an indeterminate guideline for supervising student generator $G$. 

\begin{figure}
  \centering
  \subfigure[\texttt{FFHQ}]{\includegraphics[width=0.23\textwidth]{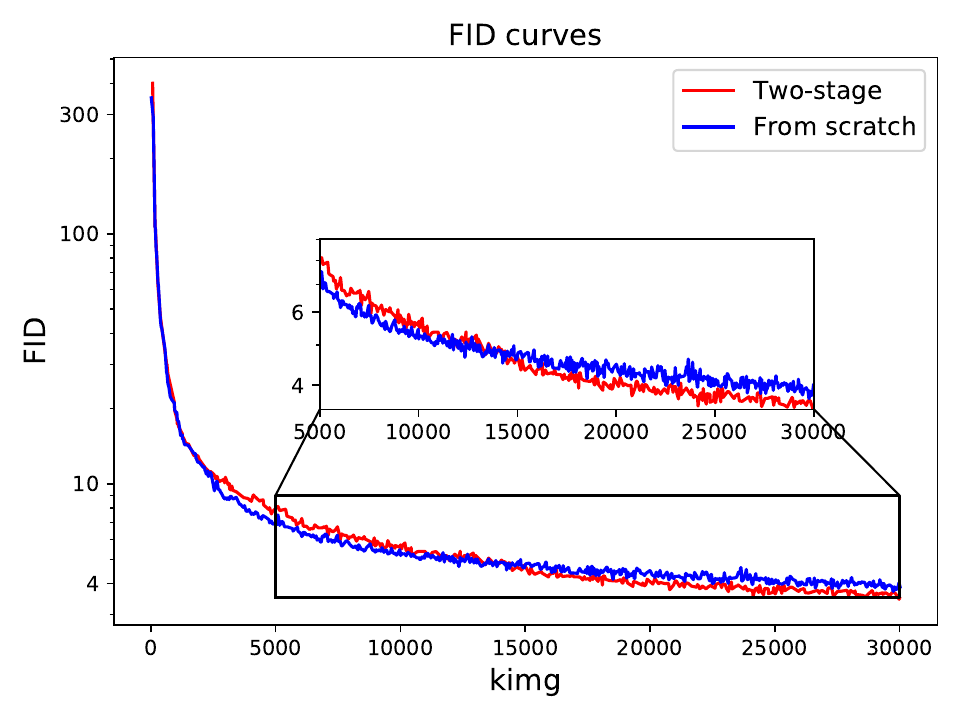}}
  \subfigure[\texttt{ImageNet}]{\includegraphics[width=0.23\textwidth]{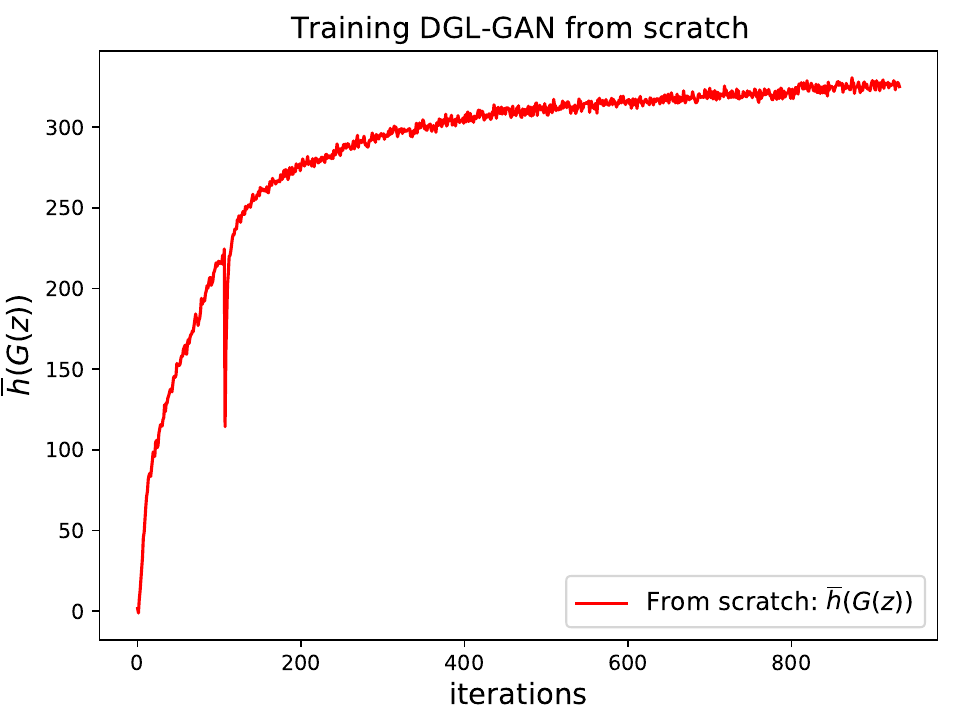}}
  \vspace{-0.3cm}
  \caption{\small (a) shows FID curves of training from scratch and two-stage training on StyleGAN2, and (b) shows the divergence behavior of training from scratch on BigGAN. Note that DGL-GAN on ImageNet exploits hinge loss, following BigGAN \cite{brock2018large}, and ${\rm Adv}(G,\overline{D})=-\overline{D}(G(z))$. Thus we directly plot the curve of $\overline{D}(G(z))$.}
  \label{Fig: From_scratch}
\end{figure}

To illuminate the impact of training DGL-GAN from scratch, we plot the curves of FID or ${\rm Adv}(G,\overline{D})$ during the optimization of formulation Eq. \eqref{full_objective} of DGL-GAN on FFHQ and ImageNet, respectively, shown in Fig. \ref{Fig: From_scratch}. We observe that the behavior on FFHQ is normal, yet the behavior on ImageNet is abnormal. On ImageNet, after a few iterations, the teacher discriminator $\overline{D}$ is fooled by the student generator $G$, coinciding with our inspection that $\overline{D}$ is not suitable for supervising $G$ at the beginning stage of training. The difference between FFHQ and ImageNet may be due to the property of the dataset. FFHQ contains a single scene, i.e., human face, while ImageNet contains many scenes, e.g., a cat and a dog, impeding the learning of the discriminator on ImageNet. Given the bounded capacity of discriminators, $\overline{D}$ of StyleGAN2 on FFHQ preserves the ability to distinguish initial fake images under the relatively simple scene. On the contrary, $\overline{D}$ of BigGAN on ImageNet cannot distinguish the initial fake images under the complex scene.

\begin{algorithm}[h!]
\small
\caption{\small Two-stage training strategy for DGL-GAN }
\label{Compressing-GANs}
{\bf Parameters:} Initialize parameters $\theta$ for the student generator. Initialize parameters $\psi$ for the student discriminator. Initialize base learning rate $\eta$, momentum parameter $\beta_1$, and exponential moving average parameter $\beta_2$ for Adam optimizer. 
\begin{algorithmic}[1] 
\State{\bf{// Stage I: training from scratch for student GAN}}
\State Set $(\theta^{1}, \psi^{1}) = (\theta, \psi)$;
\For {$t=1,2,\cdots, T$}
\State Sample real data $\{x^{(l)}\}_{l=1}^{m}\sim\mathbb{P}_{r}$ from training set and noise $\{z^{(l)}\}_{l=1}^{m}\sim\mathbb{P}_{z};$ Estimate gradient of {\rm Adv} loss in Eq.~\eqref{original function} with $\{x^{(l)},z^{(l)}\}$ at $(\theta^t,\psi^t)$, dubbed $\nabla{\rm Adv}(\theta^t,\psi^t);$ 
\State $\theta^{t+1}={\rm Adam}\big(\nabla_{\theta}{\rm Adv}(\theta^t,\psi^t),\theta,\eta,\beta_1,\beta_2\big);$
\State $\psi^{t+1}={\rm Adam}\big(\nabla_{\psi}{\rm Adv}(\theta^t,\psi^t),\psi,\eta,\beta_1,\beta_2\big);$
\EndFor 
\State{\bf{// Stage II: fine-tuning with the teacher discriminator $\overline{D}$}}
\For {$s=1,2,\cdots,S$} 
\State Sample real data $\{x^{(l)}\}_{l=1}^{m}\sim\mathbb{P}_{r}$ from the training set and latent variables $\{z^{(l)}\}_{l=1}^{m}\sim\mathbb{P}_{z}$. Estimate gradient of the DGL-GAN loss in Eq.~\eqref{full_objective} with $\{x^{(l)},z^{(l)}\}$ at $(\theta_{T+s},\psi_{T+s})$, dubbed $\nabla{V}(\theta_{T+s},\psi_{T+s});$\\
\State $\theta^{T+s+1}={\rm Adam}\big(\nabla_{\theta}V(\theta^{T+s},\psi^{T+s}),\theta,\eta,\beta_1,\beta_2\big);$
\State $\psi^{T+s+1}={\rm Adam}\big(\nabla_{\psi}V(\theta^{T+s},\psi^{T+s}),\psi,\eta,\beta_1,\beta_2\big);$
\EndFor
\State Return $(\theta,\psi)$=$(\theta^{T+S},\psi^{T+S}).$
\end{algorithmic}
\end{algorithm}

To mitigate this dilemma, we propose a two-stage training strategy for training DGL-GAN. 
In Stage I, we train vanilla GAN Eq.~\eqref{original function} with a small number of channels from scratch by discarding the term ${\rm Adv}(G,\overline{D})$ in DGL-GAN Eq.~\eqref{full_objective}. The first stage of training of DGL-GAN is as easy as training teacher GANs from scratch. 
In Stage II, we finetune DGL-GAN Eq.~\eqref{full_objective} with the supervision of teacher discriminator $\overline{D}$ on the top of the student GAN obtained in Stage I to further facilitate the performance of the student generator $G$. Thanks to the proposed two-stage training strategy, the training of Stage II is more stable than directly training DGL-GAN Eq.~\eqref{full_objective} from scratch, which also yields a better performance verified by ablation study (see Table \ref{from scratch} in section \ref{sec:ablation_study}). 
The detailed procedure of the proposed two-stage training strategy for DGL-GAN model is presented in Algorithm \ref{Compressing-GANs}.

\section{Experiments}\label{experiments-sec}

\begin{table*}
\centering
  \caption{Compressing StyleGAN2 on FFHQ. $\dagger$ denotes that StyleGAN2 trained with config-B and Fourier features in \cite{karras2021alias}.}
  \label{compress StyleGAN2 FFHQ}
  \resizebox{0.99\textwidth}{!}{
  \begin{tabular}{c|ccc|cccc}
    \toprule
    Name & \!\!Ch-Mul\!\!\!\! & \!\!\! Params (M)\!\!\!& \!\!\!MACs (G)\!\!\! & \tabincell{c}{Inference speed\\ (s)} & \tabincell{c}{PPL\\ (with central cropping)} & \tabincell{c}{PPL\\ (w/o central cropping)} & FID \\
    \midrule
    StyleGAN2 \cite{karras2020analyzing} & $1$ & \!\!\!\! $30.37$  \!\!\!\!& $142.02$ & $1.49$ & $129.4$ & $\textbf{145.0}$ & $2.84$ \\
    StyleGAN2$\dagger$ \cite{karras2021alias} & $1$ & $30.37$ & $142.02$ & $0.55$ & $\textbf{51.6}$ & $145.1$ & $\textbf{2.70}$ \\
    StyleGAN3-T \cite{karras2021alias} & $1$ & $22.31$ & $531.40$ & $0.46$ & $169.3$ & $808.7$ & $2.79$ \\
    StyleGAN3-R \cite{karras2021alias} & $1$ & $15.09$ & $235.84$ & $0.50$\tablefootnote{Note that StyleGAN3-R with minibatch $48$ occurs memory overflow. Thus we test the inference time with minibatch $24$, and report the time multiplying $2$.} & $171.4$ & $799.0$ & $3.07$ \\
    \midrule
    CA-GAN \cite{liu2021content} & $3/10$ & - & $13.39$ & $0.75$ & - & - & $7.60$ \\
    VEM \cite{kang2022information} (256$\times$256) & $3/10$ & - & $7.68$ & - & - & - & $7.48$ \\
    \!\!\!MobileStyleGAN \cite{belousov2021mobilestylegan}  \!\!\!& - & $8.01$ & $15.09$ & - & - & - & $7.75$  \!\!\!\\
    FastGAN \cite{liu2021towards} & - & \!\!\!\! $15.18$  \!\!\!\!& - & - & - & - & $12.38$  \!\!\!\\
    \multirow{3}*{Anycost-GAN \cite{lin2021anycost}} & $3/4$ & $19.39$ & $80.24$ & $1.33$ & $146.0$ & $145.5$ & $3.05$ \\
     & $1/2$ & $11.08$ & $35.61$ & $1.02$ & $127.1$ & $147.2$ & $3.28$ \\
     & $1/4$ & $5.39$ & $8.97$ & $0.43$ & $158.3$ & $158.0$ & $5.01$ \\
    \midrule
    \multirow{4}*{DGL-GAN} & $1$ & $30.37$ & $142.02$ & $1.49$ & $140.6$ & $\textbf{160.3}$ & $\textbf{2.65}$ \\
     & $1/2$ & $11.08$ & $35.61$ & $0.67$ & $\textbf{129.6}$ & $164.5$ & $2.97$ \\
     & $1/4$ & $5.39$ & $8.97$ & $0.43$ & $149.2$ & $163.1$ & $4.75$ \\
     & $1/8$ & $3.55$ & $2.28$ & $0.34$ & $228.5$ & $223.6$ & $8.13$ \\
    \bottomrule
  \end{tabular}}
\end{table*}

\begin{table}
  \caption{Compressing StyleGAN2 on LSUN-church. $\ddagger$ denotes the results we reproduced.}
  \label{compress StyleGAN2 church}
  \centering
  \begin{tabular}{c|ccc|c}
    \toprule
    Name & Ch-Mul & Params & MACs (G) & FID \\
    \midrule
    StyleGAN2 \cite{karras2020analyzing} & $1$ & $30.03$M & $83.79$ & $3.86$ \\
    StyleGAN2$\ddagger$ & $1$ & $30.03$M & $83.79$ & $\textbf{3.35}$ \\
    \midrule
    \multirow{3}*{DGL-GAN} & $1/2$ & $10.93$M & $20.99$ & $\textbf{3.78}$ \\
     & $1/4$ & $5.33$M & $5.28$ & $5.97$ \\
     & $1/8$ & $3.52$M & $1.34$ & $11.46$ \\
    \bottomrule
  \end{tabular}
\vspace{-0.2cm}
\end{table}

We verify the efficacy of the proposed DGL-GAN by applying it to compress two most representative state-of-the-art large scale vanilla GANs, i.e., StyleGAN2 \cite{karras2020analyzing} with FFHQ \cite{karras2019style} ($1024\times1024$ resolution) and LSUN-church \cite{yu2015lsun} ($256\times256$ resolution),  BigGAN \cite{brock2018large} with ImageNet \cite {deng2009imagenet} ($128\times128$ resolution). According to Algorithm \ref{Compressing-GANs}, GANs are firstly trained from scratch according to Eq. \eqref{original function}, then fine-tuned under Eq. \eqref{full_objective}. As illustrated in Section \ref{sec: Reduce width}, we compress the student generator $G$ via simply reducing the width of $G$, i.e., the intermediate features and convolution weights of $G$ are both narrowed. We use ``channel multiplier", abbreviated as \textbf{Ch-Mul}, to denote the quotient of the channels of the original generator divided by the channels of the student generator.

The section is organized as follows, Section \ref{sec: compress StyleGAN2} and Section \ref{sec: compress BigGAN} present the results of applying DGL-GAN to StyleGAN2 \cite{karras2020analyzing} and BigGAN \cite{brock2018large}, respectively. Section \ref{sec: Image Editing} presents the application of DGL-GAN on image editing. Section \ref{sec:ablation_study} conducts comprehensive ablation studies on developing conditional knowledge techniques on vanilla GANs, training uncompressed DGL-GAN, and the effectiveness of the terms in DGL-GAN, etc.

\subsection{Compressing StyleGAN2}\label{sec: compress StyleGAN2}

{\bf Implementation details}. We compress StyleGAN2 on FFHQ and LSUN-church. The channel multipliers are set as $1/2$, $1/4$, and $1/8$, respectively. On FFHQ, with channel multiplier $1/2$ and $1/4$, $T$ is set as $15000$ kimg (i.e., Stage I) and $S$ is set as $65000$ kimg (i.e., Stage II). On FFHQ, with channel multiplier $1/8$, $T$ is set as $15000$ kimg (i.e., Stage I) and $S$ is set as $45000$ kimg (i.e., Stage II). On LSUN-church, $T$ is set as $15000$ kimg and $S$ is set as $30000$ kimg, respectively. The hyperparameter $\lambda$ is set as $0.1$. Other details remain the same as StyleGAN2 \cite{karras2020analyzing}.

The results of compressed StyleGAN2 are shown in Table \ref{compress StyleGAN2 FFHQ} and Table \ref{compress StyleGAN2 church}, where we exploit commonly used FID and perceptual path length (PPL) proposed in \cite{karras2019style} to assess the quality of generated images. Regarding FFHQ, compared with original StyleGAN2 \cite{karras2020analyzing}, DGL-GAN with channel multiplier $1/2$ or nearly $1/3$ parameters obtains comparable performance on FID (2.97 vs 2.84) and PPL with central cropping (129.6 vs 129.4), demonstrating the effectiveness of training $G$ with the additional supervision of the teacher discriminator $\overline{D}$. DGL-GAN with smaller parameters (e.g., channel multiplier $1/4$ and $1/8$) also achieves acceptable performance, i.e., FID 4.75 and FID 8.13, respectively. Regarding LSUN-church, DGL-GAN obtains superior results, compared with original full StyleGAN2, DGL-GAN with channel multiplier $1/2$ also obtains comparable results (3.78 vs 3.35).

We also test the acceleration ratio of DGL-GAN on a single Nvidia Tesla V100 with 32GB memory, where models of DGL-GAN trained on FFHQ are exploited. The models are fed with latent codes with a minibatch size $48$, and the inference time is measured via taking the average over $200$ runs. To be noted, due to the narrow network of DGL-GAN, it can synthesize more images than original StyleGAN2 via a larger minibatch size, however, for a fair comparison, we still choose minibatch size $48$ for all models. Regarding inference time, compared with original StyleGAN2, DGL-GAN achieves $2.22$ acceleration ratio with channel multiplier $1/2$, $3.47$ acceleration ratio with channel multiplier $1/4$, and $4.38$ acceleration ratio with channel multiplier $1/8$. From channel multiplier $1/4$ to channel multiplier $1/8$, the acceleration ratio is not apparent, which is related to the architecture of GPU. Thus, in Section \ref{sec: Image Editing}, we exploit DGL-GAN with channel multiplier $1/4$, which achieves a trade-off between performance and efficiency. Compared with StyleGAN3 \cite{karras2021alias}, the acceleration effect of DGL-GAN is not apparent, because the implementation of StyleGAN3 exploits other acceleration tricks, e.g., FP16, etc.

As compression methods that both exploit channel multipliers, DGL-GAN outperforms Anycost-GAN under the same channel multiplier, e.g., channel multiplier $1/2$ and $1/4$. Regarding commonly-adopted FID, with channel multiplier $1/2$, DGL-GAN outperforms Anycost-GAN by $9.45\%$ (2.97 vs 3.28), and with channel multiplier $1/4$, DGL-GAN surpasses Anycost-GAN by $5.2\%$ (4.75 vs 5.01). DGL-GAN even outperforms Anycost-GAN with much higher computation complexity (i.e., 2.97 vs 3.05) on FID, manifesting the effectiveness of transferring the knowledge from the teacher discriminator $\overline{G}$ via ${\rm Adv}(G,\overline{D})$. Regarding PPL, with channel multiplier $1/2$, DGL-GAN is comparable with Anycost-GAN, i.e., 129.6 vs 127.1. Compared with recently proposed CA-GAN \cite{liu2021content}, DGL-GAN with channel multiplier $1/4$ outperforms CA-GAN by $37.5\%$ on FID (4.75 vs 7.6), where the computation complexity of DGL-GAN is about $70\%$ of CA-GAN. DGL-GAN with channel multiplier $1/4$ also surpasses MobileStyleGAN \cite{belousov2021mobilestylegan} by $61.63\%$ on FID (4.75 vs 12.38), with a much less parameter size (5.39M vs 8.01M). DGL-GAN also outperforms FastGAN \cite{liu2021towards} on FID (3.03 vs 12.38), with a more lightweight model (11.08M vs 15.18M). As for VEM \cite{kang2022information}, it only performs compression on FFHQ with the resolution of 256$\times$256, thus we cannot directly compare DGL-GAN with VEM. 



\subsection{Compressing BigGAN}\label{sec: compress BigGAN}

 {\bf Implementation details.} 
We compress BigGAN on ImageNet. The channel multipliers are set as $1/2$, $1/3$, and $1/4$, respectively. Similarly, the hyperparameter $\lambda$ is set as $0.1$. The length of stage I is set as $100000$ kimg, and stage II ends with the divergence of BigGAN, which indicates that the total iterations of training DGL-GAN on BigGAN are uncertain. Note that distinct from StyleGAN2, BigGAN shows an apparent tendency of divergence, pointed out in \cite{brock2018large}. Thus we train DGL-GAN until divergence, to explore the extreme performance DGL-GAN can achieve. We obtain pretrained $\overline{D}$ in the official \href{repository}{https://github.com/ajbrock/BigGAN-PyTorch} of BigGAN \cite{brock2018large}. Other settings remain the same as the original BigGAN.

\begin{table}
\centering
  \caption{Compressing BigGAN on ImageNet.}
  \label{Tab: compress BigGAN}
  \resizebox{0.48\textwidth}{!}{
   \begin{tabular}{c|ccc|cc}
    \toprule
    Name &\!\! Ch-Mul\!\!\! & \!\!\!\tabincell{c}{\\ (M)} \!\!\! & \tabincell{c}{MACs\\ (G)} & IS & FID \\
    \midrule
    BigGAN \cite{brock2018large} & $1$ & $70.33$ & $432.98$ & $98.8\pm0.3$ & $8.7$ \\
    DiffAug \cite{zhao2020differentiable} & $1$ & $70.33$ & $432.98$ & $100.8\pm0.2$ & $6.80$ \\
    Omni-INR-GAN \cite{zhou2021omni} & $1$ & $70.33$ & $432.98$ & $\textbf{262.85}$ & $\textbf{6.53}$ \\
    \midrule
    \!\!\!\!Slimmable GAN \cite{hou2020slimmable}\!\! & $1/2$ & $18.38$ & $211.35$ & $32.7$ & $32.8$ \\
    \midrule
    DGL-GAN & $1/2$ & $18.38$  & $211.35$ &\!\!\! $93.29\pm2.07$ & $9.92$ \\
    DGL-GAN & $1/3$ & $8.54$ & $139.76$ &\!\!\! $83.56\pm1.30$ & $12.83$ \\
    DGL-GAN & $1/4$ & $5.03$ & $104.39$ & $56.72\pm1.27$ & $19.04$ \\
    \bottomrule
  \end{tabular}}
\end{table}

The results of utilizing DGL-GAN to compress BigGAN are presented in Tab. \ref{Tab: compress BigGAN}. To be highlighted, DGL-GAN with channel multiplier $1/2$ achieves a little inferior results with original BigGAN, i.e., IS 93.29 vs 98.8 and FID 9.92 vs 8.7, illustrating that the redundancy in the networks of GANs and the effectiveness of introducing $\overline{D}$ to explore the potential in the student generator $G$. To the best of our knowledge, DGL-GAN is the first vanilla GAN compression method to achieve comparable performance with original BigGAN. DGL-GAN with channel multiplier $1/3$ reaches the slightly degenerate performance with nearly $12\%$ parameter size of the original BigGAN, i.e., IS 83.56 vs 95.08 and FID 12.83 vs 9.81. With the decrease of parameter size of DGL-GAN, we can observe that the performance drop is very sharp, especially the channel multiplier from $1/3$ to $1/4$, verifying that vanilla GANs are parameter sensitive, also illustrated in the first difficult point of compressing vanillas GANs in Section \ref{sec:introduction} and Fig. \ref{Fig: Mask Channels StyleGAN2}.

Furthermore, under the same parameters, DGL-GAN is better than Slimmable GAN \cite{hou2020slimmable} in both IS and FID. Note that the big difference between Slimmable GAN and DGL-GAN is partially due to the difference of training iterations, i.e., more than $200000$ iterations for DGL-GAN and $50000$ iterations for Slimmable GAN. However, Slimmable GAN \cite{hou2020slimmable} does not release the code for compressing BigGAN, so we cannot reproduce their results and borrow the results from their publication directly.


\subsection{Ablation Study}\label{sec:ablation_study}
In this section, we conduct extensive ablation studies to verify the effectiveness of DGL-GAN, e.g., transferring the knowledge from $\overline{D}$, and two-stage training. We also deploy other distillation techniques on StyleGAN2 to compare with DGL-GAN, e.g., traditional knowledge distillation, learning from teacher generator $\overline{G}$, gradual distillation, etc. Moreover, to further verify the effectiveness of DGL-GAN, we conduct experiments on uncompressed StyleGAN2 with the formulation of DGL-GAN.

{\bf Original Uncompressed DGL-GAN.} DGL-GAN is effective in compressed GANs, where the width of the generator and discriminator is reduced. A natural question arises, that can DGL-GAN be applied to boost the performance of original uncompressed GANs? In classification tasks, a hypothesis has been validated, that the teacher model can provide a rich source of supervision signal, leading to a student model with better generalization ability \cite{furlanello2018born}, where the teacher model and the student model share an identical architecture and parameter size. Does the hypothesis still hold for original uncompressed GANs? We conduct experiments on original uncompressed GANs with DGL-GAN.

Specifically, student models $G$ and $D$ are not compressed, sharing the identical architecture and parameter size to teacher models, which are already trained. We still adopt two-stage training strategy like DGL-GAN, to stabilize the training process. Stage I is from $0$ kimg to $20000$ kimg, and stage II is from $20000$ kimg to $30000$ kimg. The results of uncompressed DGL-GAN are presented in Tab. \ref{compress StyleGAN2 FFHQ}.

\begin{figure}[t]
  \centering
  \subfigure[\texttt{FID}]{\label{Fig: FID curve of uncompressed DGL-GAN}\includegraphics[width=0.22\textwidth]{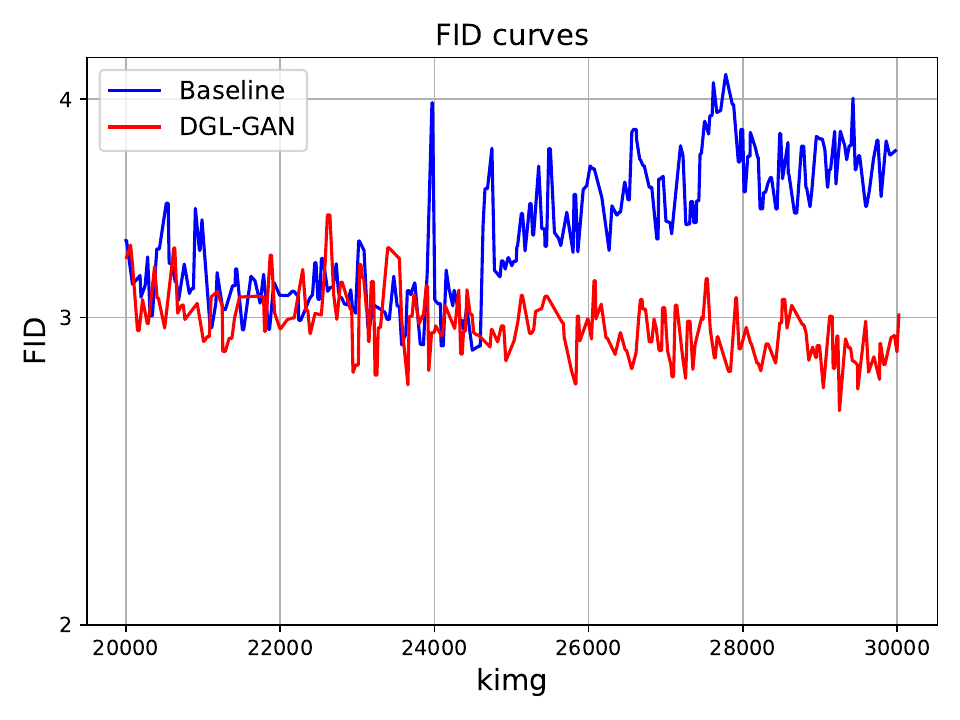}}
  \subfigure[\texttt{$h(x)$ and $h(G(z))$}]{\label{Fig: real and fake scores of uncompressed DGL-GAN}\includegraphics[width=0.22\textwidth]{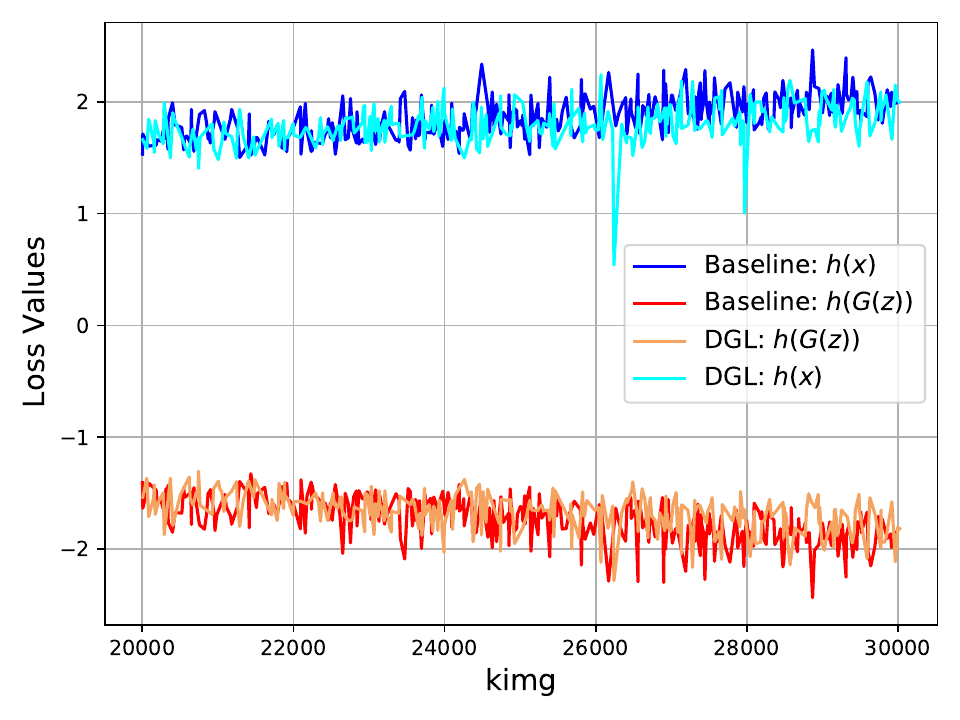}}
  \vspace{-0.35cm}
  \caption{The curves of FID and the output scores ($h(x)$ for real samples, $h(G(z))$ for fake samples) of the discriminator in the training process of uncompressed DGL-GAN and StyleGAN2, denoted as ``Baseline" in the figure. For comparison, we only plot the stage from $20000$ kimg to $30000$ kimg.}
  \label{Fig: uncompress DGL-GAN}
\end{figure}

Interestingly, uncompressed DGL-GAN outperforms StyleGAN2 by $6.7\%$ on FID, i.e., 2.65 vs 2.84, indicating that DGL-GAN works not only on compressed vanilla GANs but also on uncompressed vanilla GANs. To be highlighted, regarding FID, uncompressed DGL-GAN also outperforms StyleGAN2 with Fourier features and StyleGAN3 in \cite{karras2021alias}, achieving state-of-the-art on FFHQ only via the formulation of DGL-GAN Eq. \eqref{full_objective}. Note that the teacher discriminator $\overline{D}$ is the official released model of StyleGAN2, the companion of the pretrained generator $\overline{G}$ with FID $2.84$. However, with the combination of ${\rm Adv}(G,D)$ and ${\rm Adv}(G,\overline{D})$, the student generator $G$ even achieves better performance than the pretrained generator $\overline{G}$, which can be interpreted as that $D$ and $\overline{D}$ provide supervision signals from different perspectives, which may be beneficial to $G$. 

To better understand uncompressed DGL-GAN, we also train StyleGAN2 for 30000 kimg, similar to DGL-GAN, and we plot the curves of FID in Fig. \ref{Fig: uncompress DGL-GAN}\subref{Fig: FID curve of uncompressed DGL-GAN} and the curves of $h(x)$ and $h(G(z))$ in Fig. \ref{Fig: uncompress DGL-GAN}\subref{Fig: real and fake scores of uncompressed DGL-GAN}. We can see that when FID curve of StyleGAN2 starts to increase at about $25000$ kimg, while FID curve of DGL-GAN still keeps decreasing, indicating that DGL-GAN further explores the potential of the generator. Moreover, in Fig. \ref{Fig: uncompress DGL-GAN}\subref{Fig: real and fake scores of uncompressed DGL-GAN}, regarding the output scores $h(x)$ and $h(G(z))$ of $D$, there is no difference between DGL-GAN and StyleGAN2, illustrating that the impact of $D$ in DGL-GAN and StyleGAN2 remains identical and the performance boost of DGL-GAN is introduced by the teacher discriminator $\overline{D}$.


The superior performance of uncompressed DGL-GAN verifies the effectiveness of DGL-GAN and provides new insights about vanilla GANs. To our best knowledge, DGL-GAN is the first distillation framework to work on uncompressed GANs.

\begin{table}
  \caption{Comparison between original StyleGAN2 and DGL-GAN. Original StyleGAN2 is trained under objective function \eqref{original function}, denoted as ``Baseline" in the table.}
  \label{Tab: ablation_GDbar_stylegan2}
  \centering
  \begin{tabular}{cccc}
    \toprule
    Dataset & Name & Ch-Mul & FID \\
    \midrule
    \multirow{6}*{Church} & Baseline & $1/2$ & $4.05$ \\
    & DGL-GAN & $1/2$ & $\textbf{3.78}$ \\
    & Baseline & $1/4$ & $6.79$ \\
    & DGL-GAN & $1/4$ & $\textbf{5.97}$ \\
    & Baseline & $1/8$ & $12.52$ \\
    & DGL-GAN & $1/8$ & $\textbf{11.46}$ \\
    \bottomrule
  \end{tabular}
\vspace{-0.2cm}
\end{table}
\begin{table}
\centering
  \caption{Comparison between original BigGAN and DGL-GAN. Original BigGAN is trained under objective function \eqref{original function}, denoted as ``Baseline" in the table.}
  \label{Tab: ablation_GDbar_BigGAN}
   \begin{tabular}{cccc}
    \toprule
    Name & Ch-Mul & IS & FID \\
    \midrule
    Baseline & \multirow{2}*{$1/2$} & $88.24\pm1.42$ & $10.86$ \\
    DGL-GAN &  & $\textbf{93.29}\pm\textbf{2.07}$ & $\textbf{9.92}$ \\
    \midrule
    Baseline & \multirow{2}*{$1/3$} & $80.23\pm2.31$ & $13.34$ \\
    DGL-GAN &  & $\textbf{83.56}\pm\textbf{1.30}$ & $\textbf{12.83}$ \\
    \midrule
    Baseline & \multirow{2}*{$1/4$} & $43.95\pm0.75$ & $20.64$ \\
    DGL-GAN &  & $\textbf{56.72}\pm\textbf{1.27}$ & $\textbf{19.04}$ \\
    \bottomrule
  \end{tabular}
\end{table}

{\bf The effectiveness of ${\rm Adv}(G,\overline{D})$.} Compared with original vanilla GANs, the formulation of DGL-GAN only adds a regularization term, i.e., ${\rm Adv}(G,\overline{D})$. To verify the effectiveness of the term ${\rm Adv}(G,\overline{D})$, we train StyleGAN2 and BigGAN with same channel multipliers under original objective function \eqref{original function}, denoted as ``Baseline", presented in Tab. \ref{Tab: ablation_GDbar_stylegan2} and Tab. \ref{Tab: ablation_GDbar_BigGAN}. For fair comparison, the ``Baseline" of StyleGAN2 are trained with similar kimg to DGL-GAN and the ``Baseline" of BigGAN are also trained until divergence\footnote{For StyleGAN2, training iterations of ``Baseline" and DGL-GAN are identical. For BigGAN, training iterations of ``"Baseline" and DGL-GAN may be different, but both are trained until divergence.}. 

The results of ``Baseline" and DGL-GAN are shown in Tab. \ref{Tab: ablation_GDbar_stylegan2}, and Tab. \ref{Tab: ablation_GDbar_BigGAN}. Regarding both StyleGAN2 and BigGAN, DGL-GAN consistently outperforms ``Baseline" on FID, demonstrating the promotion impact of ${\rm Adv}(G,\overline{D})$ on the student generator $G$ with reduced width and verifying our motivation that the teacher discriminator $\overline{D}$ contains some useful information that can boost the performance of $G$, shown in Fig. \ref{Fig: The gradients of D and Dbar}.

\begin{table}[t]
\centering
  \caption{The results of training the student StyleGAN2 only with ${\rm Adv}(G,\overline{D})$ on FFHQ. ``-" denotes that the training process diverges.}
  \label{Tab: single_G_Dbar}
  \resizebox{0.48\textwidth}{!}{
   \begin{tabular}{c|ccc|c}
    \toprule
    Name &\!\! Ch-Mul\!\!\! & kimg & Strategy & FID \\
    \midrule
    DGL-GAN & $1/2$ & $30000$ & fine-tuning & $\textbf{3.37}$ \\
    Single fixed $\overline{D}$ & $1/2$ & $30000$ & fine-tuning & $4.34$ \\
    Single fixed $\overline{D}$ & $1/2$ & $30000$ & from scratch & - \\
    Single updated $\overline{D}$ & $1/2$ & $30000$ & fine-tuning & $3.47$ \\
    Single updated $\overline{D}$ & $1/2$ & $30000$ & from scratch & $6.76$ \\
    \bottomrule
  \end{tabular}}
\end{table}

{\bf ${\rm Adv}(G,D)$ is also indispensable.} To validate whether training compressed GANs under the single term ${\rm Adv}(G,\overline{D})$ is feasible, we experiment with some variants of DGL-GAN. We train the student generator $G$ with channel multiplier $1/2$ on StyleGAN2 on FFHQ, with frozen teacher discriminator $\overline{D}$ or updated teacher discriminator $\overline{D}$, abbreviated as ``Single fixed $\overline{D}$" and ``Single updated $\overline{D}$", respectively. With frozen $\overline{D}$, the objective function is
\begin{align}
    & \min_{G} {\rm Adv}(G,\overline{D}).
\end{align}
With updated $\overline{D}$, the objective function is
\begin{align}
    & \min_{G} \max_{\overline{D}} {\rm Adv}(G,\overline{D}).
\end{align}
We totally experiment with two training strategies, i.e., from scratch and fine-tuning, where stage I is $15000$ kimg and stage II is $15000$ kimg. The results are presented in Tab. \ref{Tab: single_G_Dbar}, where DGL-GAN shows the best performance among these settings, indicating that the original objective function \eqref{original function} of vanilla GANs are also vital and training only with ${\rm Adv}(G,\overline{D})$ is not optimal. The adversarial interaction between the student generator $G$ and the student discriminator $D$ is essential for the final performance, thus the combination of ${\rm Adv}(G,D)$ and ${\rm Adv}(G,\overline{D})$ is essential.

\begin{figure}[t]
  \centering
  \subfigure[\texttt{Ch-Mul $1/2$}]{\includegraphics[width=0.22\textwidth]{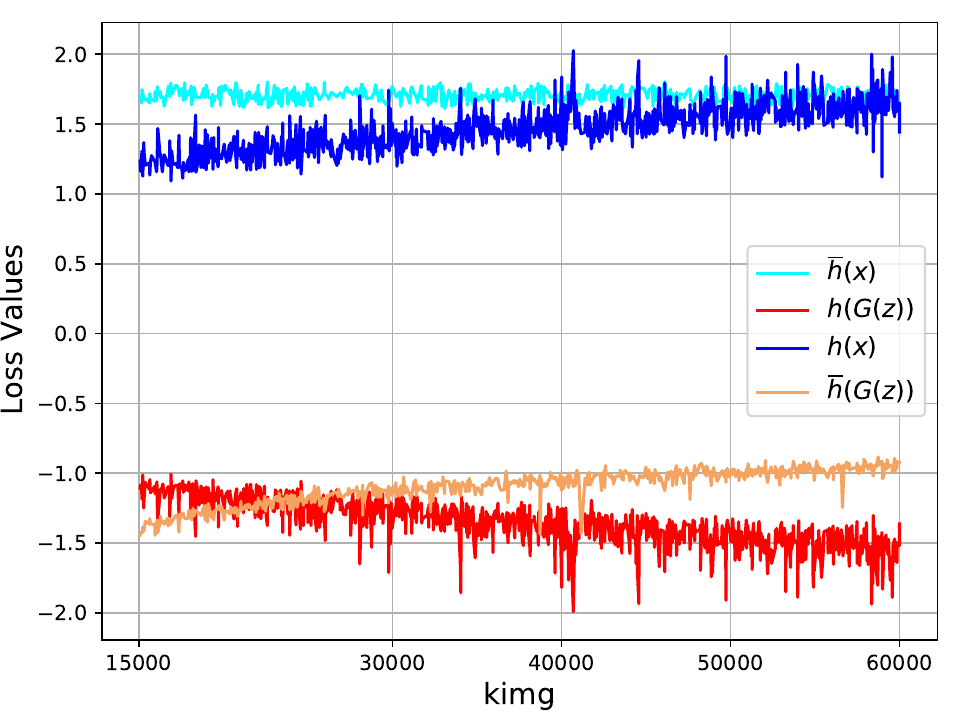}}
  \subfigure[\texttt{Ch-Mul $1/4$}]{\includegraphics[width=0.22\textwidth]{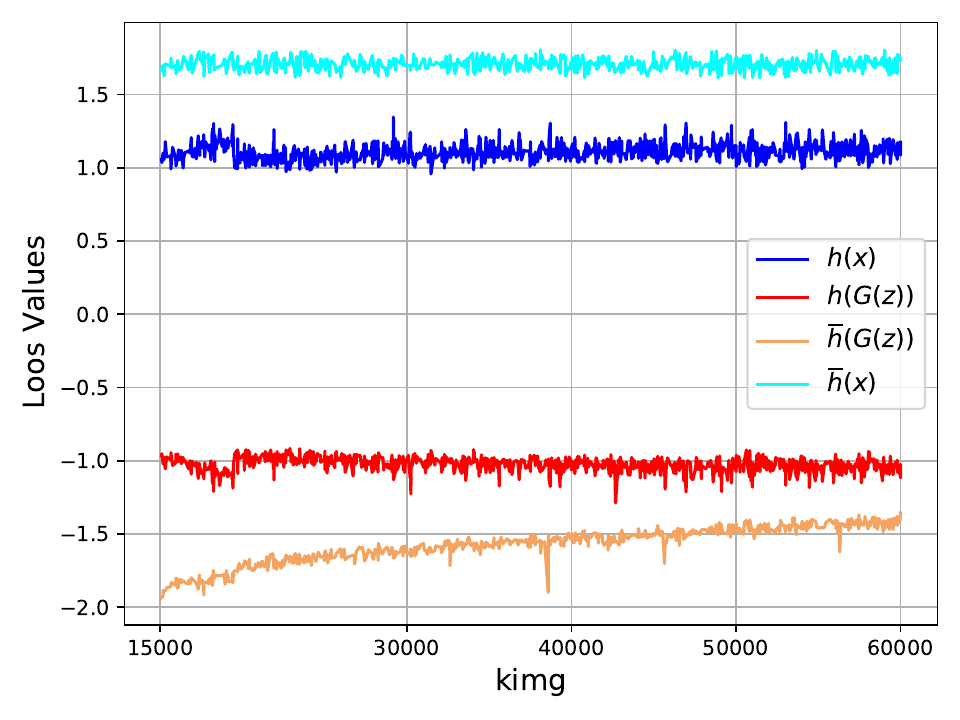}}
  \vspace{-0.35cm}
  \caption{The curves of $h(G(z))$ and $\overline{h}(G(z))$ in the stage II of DGL-GAN with channel multiplier $1/2$ and $1/4$.}
  \label{Fig: h(G(z)) and hbar(G(z)) in stage II.}
\end{figure}

{\bf Imitate teacher generator/discriminator.} In traditional knowledge distillation, the student network is optimized to imitate the intermediate features \cite{romero2014fitnets} or the final logits \cite{hinton2015distilling} of the teacher network. Previous works \cite{aguinaldo2019compressing, liu2021content, chen2021gans} on the compression of GANs, the student generator/discriminator is optimized to imitate the teacher generator/discriminator in the feature map space, shown in the following equations
\begin{align}
& \small \mathcal{L}_{inter}(G,\overline{G})=\displaystyle{\sum}_{l\in\mathcal{S}_{KD}}dist(r(G^{(l)}),\overline{G}^{(l)})\label{Eq: G_inter},\\
& \small \mathcal{L}_{inter}(D,\overline{D})=\displaystyle{\sum}_{l\in\mathcal{S}_{KD}}dist(r(D^{(l)}),\overline{D}^{(l)})\label{Eq: D_inter},\\
& \small \mathcal{L}_{out}(D,\overline{D})=CE(D,\overline{D}),
\end{align}

\begin{table}
\centering
  \caption{The results of ``G-inter", ``D-inter", and ``D-out" on FFHQ. ``out" denotes that we also align the output image of the student generator G with that of the teacher generator $\overline{G}$.}
  \label{G_L1_result}
  \begin{tabular}{c|cc|c}
    \toprule
    Name & Feature set $S_{KD}$ & kimg & FID \\
    \hline\hline
    DGL-GAN & - & $30000$ & ${\bf 3.37}$ \\
    \midrule
    G-inter & $\{256,512,1024\}$ & \multirow{6}*{$30000$} & $4.05$ \\
    G-inter & $\{256,512,1024,{\rm out}\}$ &  & $4.06$ \\
    G-inter & $\{8,32,128\}$ &  & $4.20$ \\
    G-inter & $\{8,32,128,{\rm out}\}$ &  & $4.25$ \\
    G-inter & $\{8,128,1024\}$ &  & $4.02$ \\
    G-inter & $\{8,128,1024,{\rm out}\}$ &  & $4.08$ \\
    \midrule
    D-inter & $\{4,16,64\}$ & \multirow{4}*{$30000$} & $4.18$ \\
    D-inter & $\{256,512,1024\}$ &  & $4.02$ \\
    D-inter & $\{8,128,1024\}$ &  & $4.12$ \\
    D-out & - &  & $4.40$ \\
    \bottomrule
  \end{tabular}
\vspace{-0.4cm}
\end{table}

where $\mathcal{S}_{KD}$ denotes the set of layers to be imitated by the student generator or the student discriminator, and $dist()$ is a function to measure the distance of feature maps between the student generator and teacher generator, which can be $\ell_1$ distance or $\ell_2$ distance. $r$ in Eq. \eqref{Eq: G_inter} and Eq. \eqref{Eq: D_inter} represents the 1x1 convolution operation to transform the shape of the features in $G$ or $D$. $CE$ in Eq. \eqref{Eq: D_out} denotes the cross entropy between the output distribution of $D$ and that of $\overline{D}$ on real samples and fake samples. We add the above objective functions into the original objective function Eq.~\eqref{original function}, yielding
\begin{align}\label{full_G_L1}
& \small {\min}_{G}{\max}_{D} {\rm Adv}(G,D)+\gamma_{1}*\mathcal{L}_{inter}(G,\overline{G}),\\
& \small {\min}_{G}{\max}_{D} {\rm Adv}(G,D)+\gamma_{2}*\mathcal{L}_{inter}(D,\overline{D}),\\
& \small {\min}_{G}{\max}_{D} {\rm Adv}(G,D)+\gamma_{3}*\mathcal{L}_{out}(D,\overline{D})\label{Eq: D_out},
\end{align}
which is dubbed ``G-inter", ``D-inter", and ``D-out", respectively. We tune the hyperparameters $\gamma_1$, $\gamma_2$, and $\gamma_3$, to explore the extreme performance with traditional knowledge distillation techniques. We experiment with various $S_{KD}$, i.e., different combinations of distilled layers. In the implementation, we exploit $\ell_1$ distance as $\mathcal{L}_{inter}$ and channel multiplier as $1/2$. The results of imitating teacher generator/discriminator in the feature map space or the output space are presented in Table \ref{G_L1_result}.

\begin{figure}[t]
  \centering
  \subfigure[\texttt{$\mathcal{L}_{inter}(G,\overline{G})$}]{\includegraphics[width=0.23\textwidth]{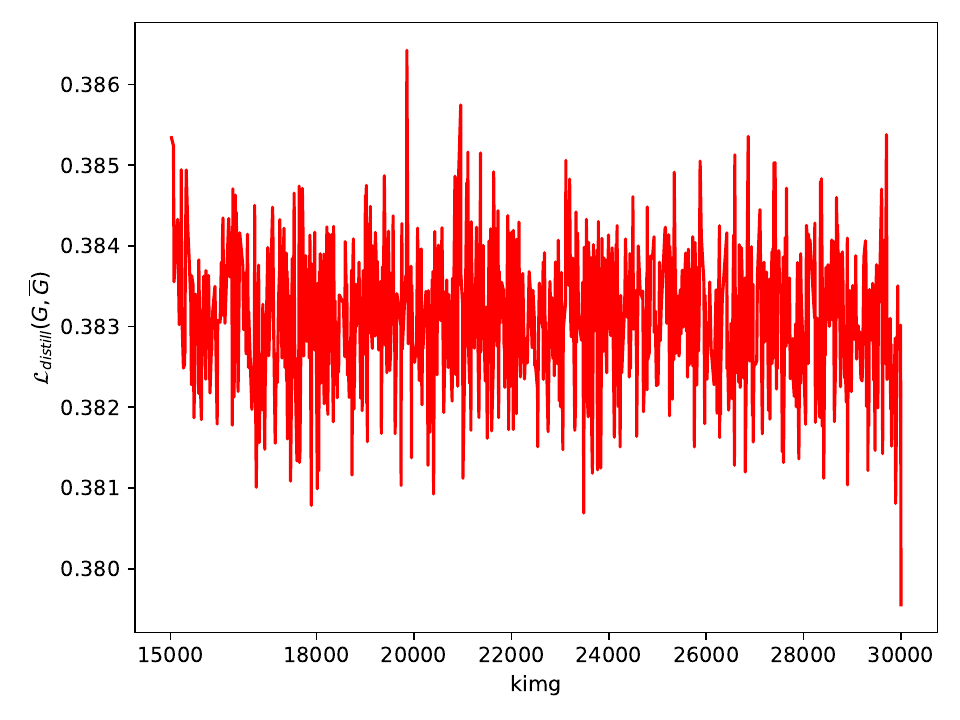}}
  \subfigure[\texttt{$\mathcal{L}_{inter}(D,\overline{D})$}]{\includegraphics[width=0.23\textwidth]{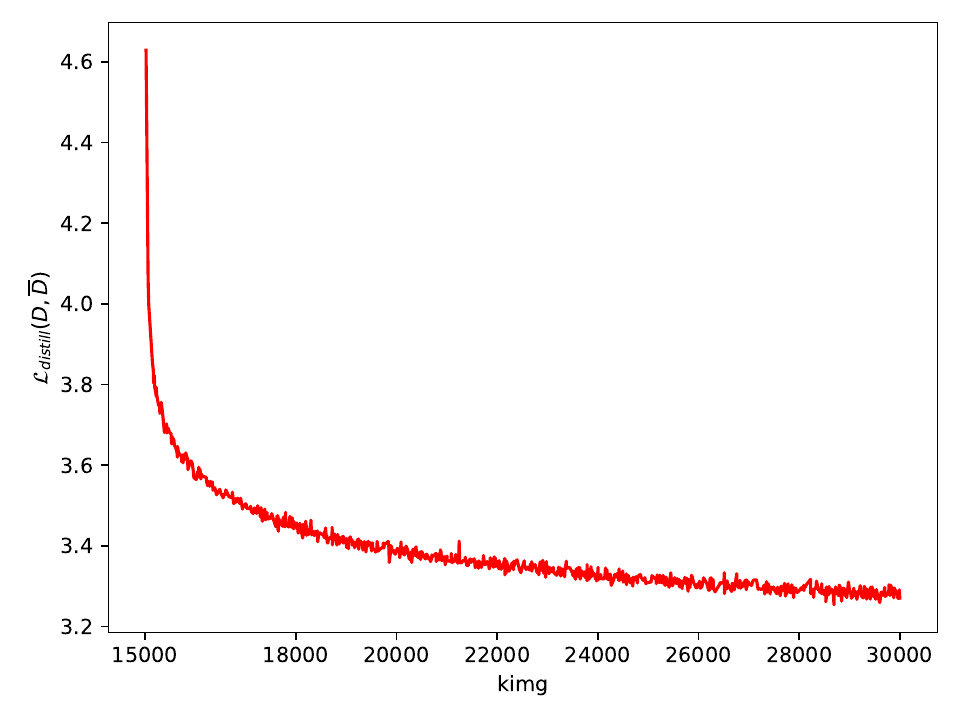}}
  \vspace{-0.35cm}
  \caption{The curves of $\mathcal{L}_{inter}(G,\overline{G})$ and $\mathcal{L}_{inter}(D,\overline{D})$ during training. $\gamma_1$ and $\gamma_2$ are both set as $0.2$. The distilled features of $G$ and $D$ are both in the top of the networks.}
  \label{Fig: G_L1 and D_L1}
\end{figure}

By Table \ref{G_L1_result}, directly exploiting ``G-inter", ``D-inter", and ``D-out" to distill the knowledge of teacher StyleGAN2 into student StyleGAN2 on FFHQ is inferior to DGL-GAN and ``Baseline" with the same channel multiplier $1/2$, illustrating that simply enforcing the student generator/discriminator to imitate the teacher generator/discriminator in the feature space or the output space is not feasible.

We analyze the phenomenon that traditional knowledge distillation techniques do not work on vanilla GANs from two perspectives. First, regarding ``G-inter", as we illustrated in the difficulty b in Section \ref{sec:introduction}, the individual input of the generator lacks ground truth. As an unsupervised task, the ground truth for the generator is the \textbf{distribution} of a real dataset, while for an individual input of the generator, there is no definition of ground truth. Given a sampled latent code $z$ to be sent to the generator, it is not essential for a generator to produce a certain image, e.g., a young girl. Observing the formulation of ``G-inter", with the same latent code $z$, minimizing the distance of intermediate features between $G$ and $\overline{G}$ sets a ground truth for $G$, which is meaningless. Thus, for noise-to-image vanilla GANs, optimizing the student generator $G$ towards imitating the teacher generator $\overline{G}$ is infeasible. Second, regarding ``D-inter" and ``D-out", they transfer the knowledge from $\overline{D}$ to $D$. However, compared with DGL-GAN, which directly transfers the knowledge from $\overline{D}$ to $G$ via term ${\rm Adv}(G,\overline{D})$, ``D-inter" and ``D-out" are less straight forward, i.e., $\overline{D}\rightarrow D\rightarrow G$ vs $\overline{D}\rightarrow G$. Thus, ``D-inter" and ``D-out" are inferior to DGL-GAN.

Further, we plot the curves of $\mathcal{L}_{inter}(G,\overline{G})$ and $\mathcal{L}_{inter}(D,\overline{D})$ in Fig. \ref{Fig: G_L1 and D_L1}. Obviously, in contrast to $\mathcal{L}_{inter}(D,\overline{D})$, $\mathcal{L}_{inter}(G,\overline{G})$ hardly decreases, indicating the difficulty of optimizing $\mathcal{L}_{inter}(G,\overline{G})$ and coinciding with our inspection that the generator lacks ground truth for an individual input.

{\bf Learn from $\overline{G}$ via adversarial objective function.} To verify that only transferring the knowledge of the teacher discriminator $\overline{D}$ via ${\rm Adv}(G,\overline{D})$ is effective, we design two alternative models. First, we transfer the knowledge of the teacher generator $\overline{G}$ via the adversarial objective function, shown as
\begin{align}\label{Eq: full_objective_generator}
    & \min_{G} \max_{D} {\rm Adv}(G,D)+\lambda{\rm Adv}(\overline{G},D),
\end{align}
which is denoted as {\bf G}enerator {\bf G}uided {\bf L}earning for GAN compression (GGL-GAN). 

In addition, we transfer the knowledge of both teacher generator $\overline{G}$ and teacher discriminator $\overline{D}$ to student GAN simultaneously as
\begin{align}\label{compress_full_GbarD_DbarG}
\small\!\!\! \min_{G} \max_{D} {\rm Adv}(G,D)\!+\!\lambda{\rm Adv}(\overline{G},D)\!+\!\lambda{\rm Adv}(G,\overline{D}), 
\end{align}
which is denoted as {\bf G}enerator {\bf D}iscriminator {\bf G}uided {\bf L}earning for GAN compression (GDGL-GAN).

\begin{table}
\centering
  \caption{\small Ablation study of transferring $\overline{G}$ or $\overline{D}$.}
  \label{G-Ad_result}
  \begin{tabular}{c|ccc|ccc|c}
    \toprule
    Name & Ch-Mul & Params & kimg & FID \\
    \hline\hline
    Baseline & \multirow{4}*{$1/2$} & \multirow{4}*{$11.08$M} & \multirow{4}*{$30000$} & $3.63$ \\
    DGL-GAN &  &  &  & ${\bf 3.37}$ \\
    GGL-GAN &  &  &  & $3.99$ \\
    GDGL-GAN &  &  &  & $3.62$ \\
    \bottomrule
  \end{tabular}
\vspace{-0.4cm}
\end{table}

We train the student StyleGAN2 on FFHQ with channel multiplier $1/2$ via the objective function Eq.~\eqref{Eq: full_objective_generator} or Eq.~\eqref{compress_full_GbarD_DbarG} with the similar training techniques for DGL-GAN and report the results in Table \ref{G-Ad_result}. In Table \ref{G-Ad_result}, both GGL-GAN and GDGL-GAN show inferior results, compared with DGL-GAN. The performance of GGL-GAN is inferior to those of ``Baseline" and GDGL-GAN, demonstrating that the regularization term ${\rm Adv}(\overline{G},D)$ hampers the performance of student GANs and the regularization term ${\rm Adv}(G,\overline{D})$ facilitates the performance of student GANs. Overall, DGL-GAN shows the best performance over three settings of compression methods.

In GGL-GAN, the optimal formulation $D*$ is affected by the term ${\rm Adv}(\overline{G},D)$, changing from $D*=\frac{p_d}{p_d+p_g}$ to $ D*={p_{d}}/\big[{p_{d}+p_g+\lambda \overline{p_g}}\big]$. Thus the knowledge of the teacher generator $\overline{G}$ is not transferred properly, and the term ${\rm Adv}(\overline{G},D)$ even hurts the performance of the student generator $G$. More details can be found in the supplementary material.

{\bf Training from scratch.} In Algorithm \ref{Compressing-GANs},  student GANs are firstly trained under the original objective function Eq.~\eqref{original function} and then trained under DGL-GAN objective function Eq.~\eqref{full_objective}, which is viewed as a two-stage strategy to find a good initial point for the optimization of DGL-GAN. To validate the effectiveness of Algorithm \ref{Compressing-GANs}, we train DGL-GAN from scratch on FFHQ, denoted as ``from scratch".

\begin{table}
\centering
  \caption{\small The ablation study of ``from scratch".}
  \label{from scratch}
  \begin{tabular}{c|ccc|ccc|c}
    \toprule
    Name & Ch-Mul & Params & kimg & FID \\
    \midrule
    DGL-GAN & \multirow{2}*{$1/2$} & \multirow{2}*{$11.08$M} & \multirow{2}*{$30000$} & ${\bf 3.37}$ \\
    from scratch &  &  &  & $3.72$ \\
    \bottomrule
  \end{tabular}
\vspace{-0.2cm}
\end{table}

The results of ``from scratch" are presented in Table \ref{from scratch}. The performance of ``from scratch" is inferior to that of DGL-GAN, illustrating the importance of a good initial point for training DGL-GAN. On the other hand, the teacher discriminator $\overline{D}$ aims to distinguish two distributions with low divergence. i.e., $p_{\rm data}$ and $\overline{p_g}$. The fake data distribution $p_{g}$ in the beginning stage of training is far from $p_{data}$. It is meaningless to exploit $\overline{D}$ to distinguish $p_{g}$ in the beginning.


Overall, we exploit the two-stage training strategy in Algorithm \ref{Compressing-GANs} for two reasons. First, the two-stage training strategy can find a good initial point for DGL-GAN and facilitate the performance of $G$. Second, the unstable training behavior can be circumvented with two-stage training behavior.

\begin{table}
\centering
  \caption{The results of gradual distillation with channel multiplier $1/4$.}
  \vspace{-0.2cm}
  \label{Tab: gradual distillation}
  \begin{tabular}{c|cc|c}
    \toprule
    Name & CH-Mul & kimg & FID \\
    \midrule
    Baseline & \multirow{3}*{$1/4$} & \multirow{3}*{$60000$} & $5.20$ \\
    DGL-GAN &  &  & $4.95$ \\
    DGL-GAN + gradual &  &  & $5.21$ \\
    \bottomrule
  \end{tabular}
\vspace{-0.4cm}
\end{table}

{\bf Gradual distillation.} Slimmable GAN \cite{hou2020slimmable} exploits gradual distillation to transfer the knowledge from teacher models to student models gradually. To validate gradual distillation on DGL-GAN, we conduct experiments with channel multiplier $1/4$, where $\overline{D}$ is the pretrained discriminator $D$ in DGL-GAN with channel multiplier $1/2$, instead of the pretrained discriminator in original StyleGAN2. Thus, the knowledge is gradually transferred, i.e., original StyleGAN2 $\rightarrow$ DGL-GAN with channel multiplier $1/2$ $\rightarrow$ DGL-GAN with channel multiplier $1/4$. The results of gradual distillation are reported in Tab. \ref{Tab: gradual distillation}.

In Tab. \ref{Tab: gradual distillation}, DGL-GAN with gradual distillation shows inferior performance compared with DGL-GAN and ``Baseline", demonstrating that the formulation of DGL-GAN requires a teacher discriminator $\overline{D}$ with strong capacity (e.g., the discriminator in the original StyleGAN2), rather than an ``intermediate discriminator".

\begin{figure}[t]
  \centering
  \includegraphics[width=0.49\textwidth]{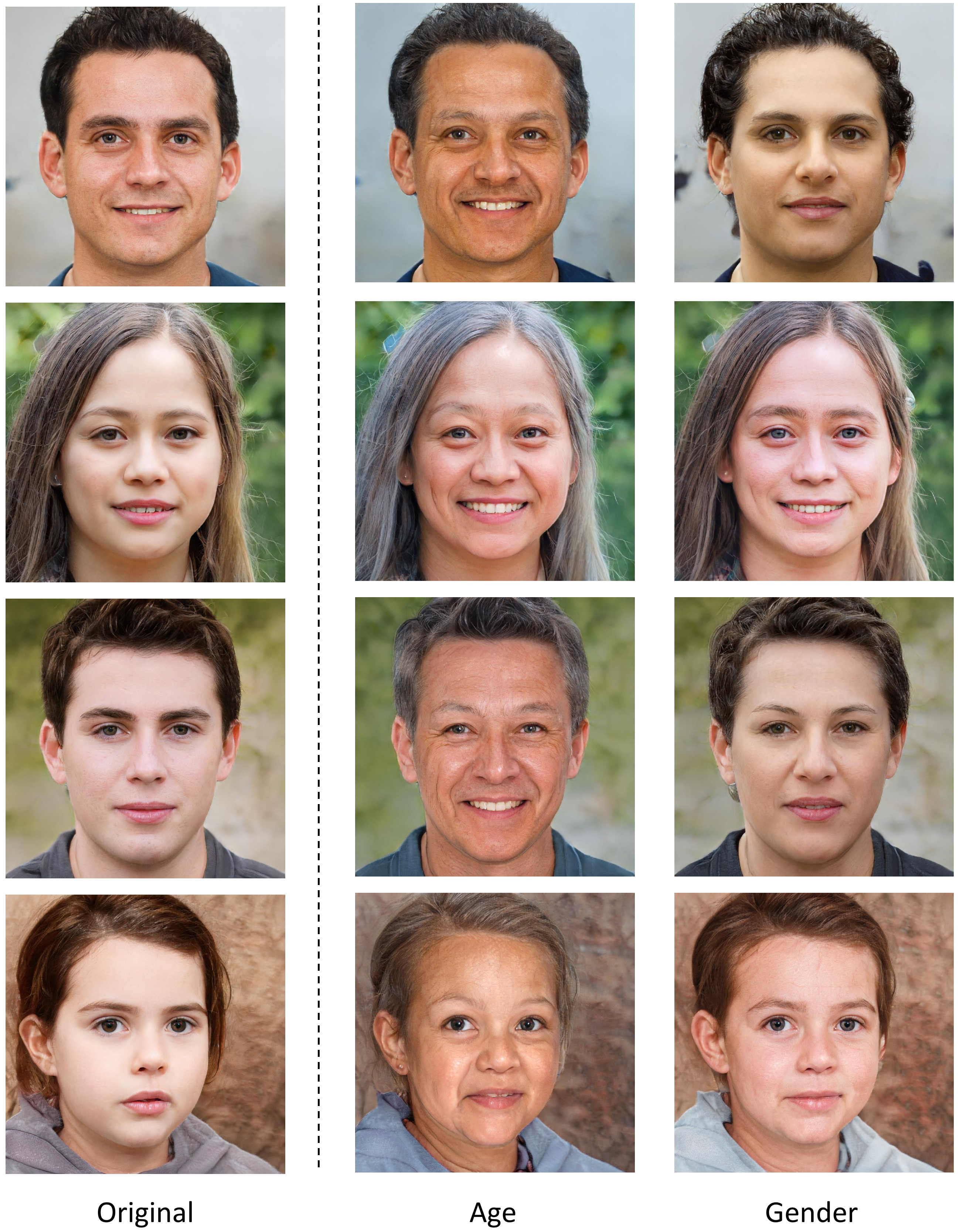}
  \vspace{-0.35cm}
  \caption{\small The edited images synthesized via DGL-GAN with channel multiplier $1/4$.}
  \vspace{-0.4cm}
  \label{Fig: Face_edit}
\end{figure}

\subsection{Image Editing}\label{sec: Image Editing}
To illustrate the quality of images generated via DGL-GAN, exploiting InterfaceGAN \cite{shen2020interfacegan}, we edit the images synthesized by the compressed model of DGL-GAN. We utilize the compressed StyleGAN2 with channel multiplier $1/4$ with FID $4.75$ on FFHQ and edit the synthesized images on age and gender properties. Latent space $\mathcal{W}$ is explored to edit images on age and gender. The images after editing are presented in Fig. \ref{Fig: Face_edit}. 
It can be observed in Fig. \ref{Fig: Face_edit} that the edited image is distinct from the original image only on the property to be edited, e.g., age or gender, indicating the quality of synthesized images of DGL-GAN is acceptable and the smoothness in the $\mathcal{W}$ latent space in the compressed model of DGL-GAN. The qualitative results manifest the potential of deploying lightweight models of DGL-GAN on image editing. 
\section{Conclusions}

DGL-GAN, a simple yet effective framework for compressing vanilla GANs, can facilitate the performance of narrow GANs via transferring the knowledge from the teacher discriminator via the adversarial objective function. To be highlighted, not confined to compressed GANs, DGL-GAN is also effective on uncompressed GANs, e.g., uncompressed StyleGAN2, and achieves state-of-the-art on a large-scale dataset FFHQ, which provides some insights about training GANs. There exists some directions to further explore. First, to further explore the impact of DGL-GAN on uncompressed GANs and analyze the reason that DGL-GAN can boost uncompressed GANs. Second, trying to obtain a more precise $\overline{D}$. Now we only exploit the pretrained discriminator or update the teacher discriminator simultaneously, with the assumption that it is closer to the global optimal discriminator. Yet can we explore a method that can obtain a more precise discriminator? Third, introducing an automatic mechanism, which can automatically determine whether to utilize the formulation of DGL-GAN, i.e., automatically setting $T$ and $S$ in Algorithm \ref{Compressing-GANs}. 
\section{Compliance with Ethical Standards}
The research does not involve Human Participants or Animals. There is no conflict of interest to report.

\backmatter





\bmhead{Data availability
}
The datasets used in this research can be found publicly. ImageNet is in this \hyperlink{https://www.image-net.org/download.php}{link}. LSUN is in this \hyperlink{https://www.yf.io/p/lsun}{link}. FFHQ is in this \hyperlink{https://drive.google.com/drive/folders/1tZUcXDBeOibC6jcMCtgRRz67pzrAHeHL}{link}.

\bmhead{Acknowledgments}
This work is supported in part by the Fundamental
Research Funds for the Central Universities. This work is also supported by Science and Technology Innovation 2030 –“Brain Science and Brain-like Research” Major Project (No. 2021ZD0201402 and No. 2021ZD0201405).










\begin{appendices}

\section{Experiment Details}\label{experiment_details}
In this section, we provide the details of our experiments. All devices exploited in our experiments are Tesla V100 GPU with 32GB memory. The metric FID we exploit calculates the distance between $50$k images produced by the generator and real images, same as StyleGAN2 and BigGAN.

{\bf Compressing StyleGAN2.}
FFHQ dataset contains $70000$ images, with $1024\times1024$ resolution. LSUN-church dataset contains $126227$ images, with $256\times256$ resolution in training. FID is measured via calculating the distance between $50000$ generated images and images in FFHQ dataset or LSUN-church dataset, respectively. On FFHQ and LSUN-church, the learning rate is set to $0.002$ for both the generator and the discriminator, and $\beta_{0}$ and $\beta_{1}$ of Adam optimizer are set to $0.0$ and $0.99$, respectively. The minibatch size of FFHQ and LSUN-church is $32$. On FFHQ, the compressed StyleGAN2 is trained on 4 GPU devices. On  LSUN-church, compressed StyleGAN2 is trained on 4 GPU devices. Other settings of training remain the same as original StyleGAN2. For FFHQ, we exploit the teacher discriminator $\overline{D}$ released by the authors of StyleGAN2 \cite{karras2020analyzing}. For LSUN-church, we exploit the teacher discriminator $\overline{D}$ trained by us, of which FID is $3.35$.

{\bf Compressing BigGAN.}
ImageNet contains $1281167$ images in the training dataset, $50000$ images in the validation dataset, and $150000$ images in the test dataset. We use the training dataset to compress BigGAN. FID is measured via calculating the distance between $50000$ generated images and images in the training dataset, where Inception-v3 network in PyTorch is exploited. Specifically, for the two-stage training strategy, $T$ is set to $100000$ and $S$ is uncertain. We train stage-II of DGL-GAN until the model diverges, as well as ``Baseline". During the training process, the minibatch size is $256$, while the other settings remain the same as original BigGAN \cite{brock2018large}. The learning rate of the generator is set to $0.0001$, and the learning rate of the discriminator is set to $0.0004$. $\beta_0$ and $\beta_1$ are set to $0.0$ and $0.999$, respectively. As for the teacher discriminator $\overline{D}$, we exploit the discriminator model released by the authors of BigGAN \cite{brock2018large}.

\begin{figure*}[t]
  \vspace{-0.5cm}
  \centering
  \subfigure[\texttt{seed2004\_orginal}]{\includegraphics[width=0.48\textwidth]{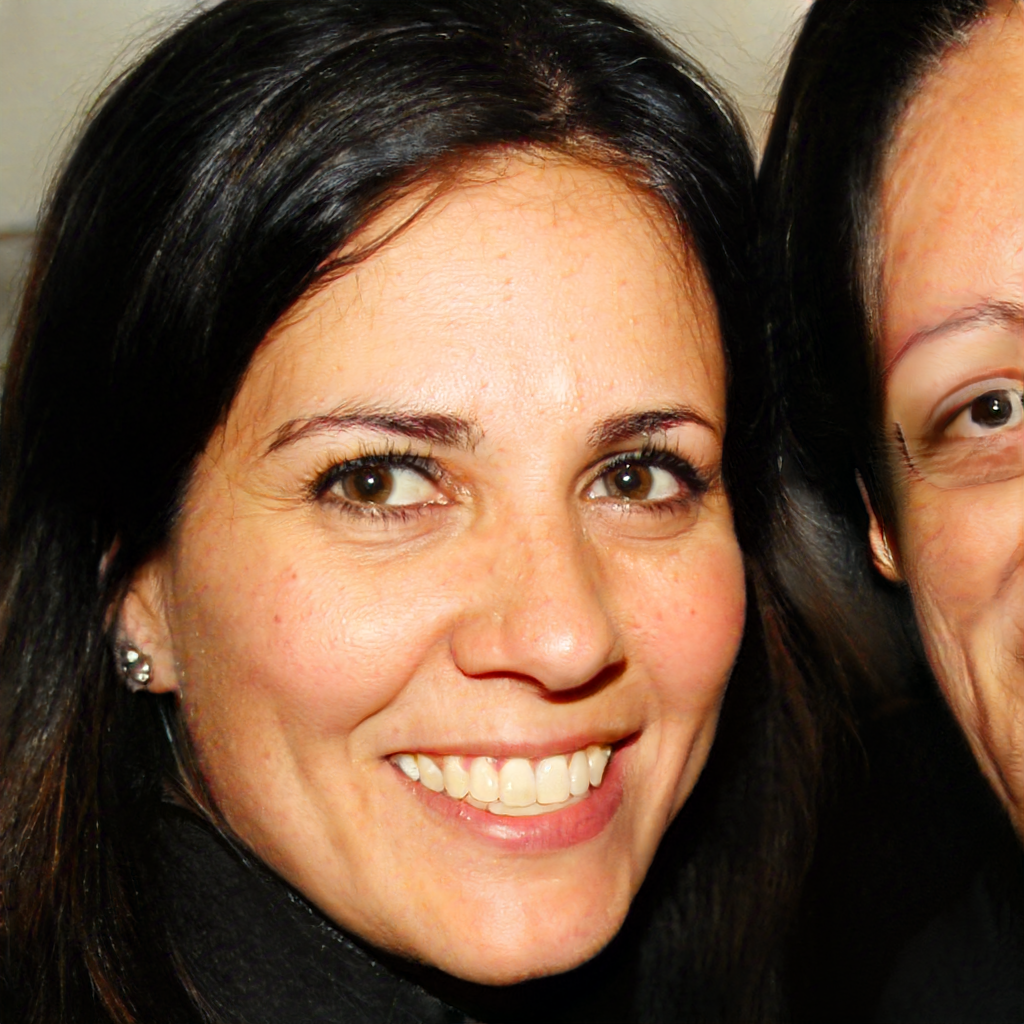}}
  \subfigure[\texttt{seed2004\_DGL-GAN}]{\includegraphics[width=0.48\textwidth]{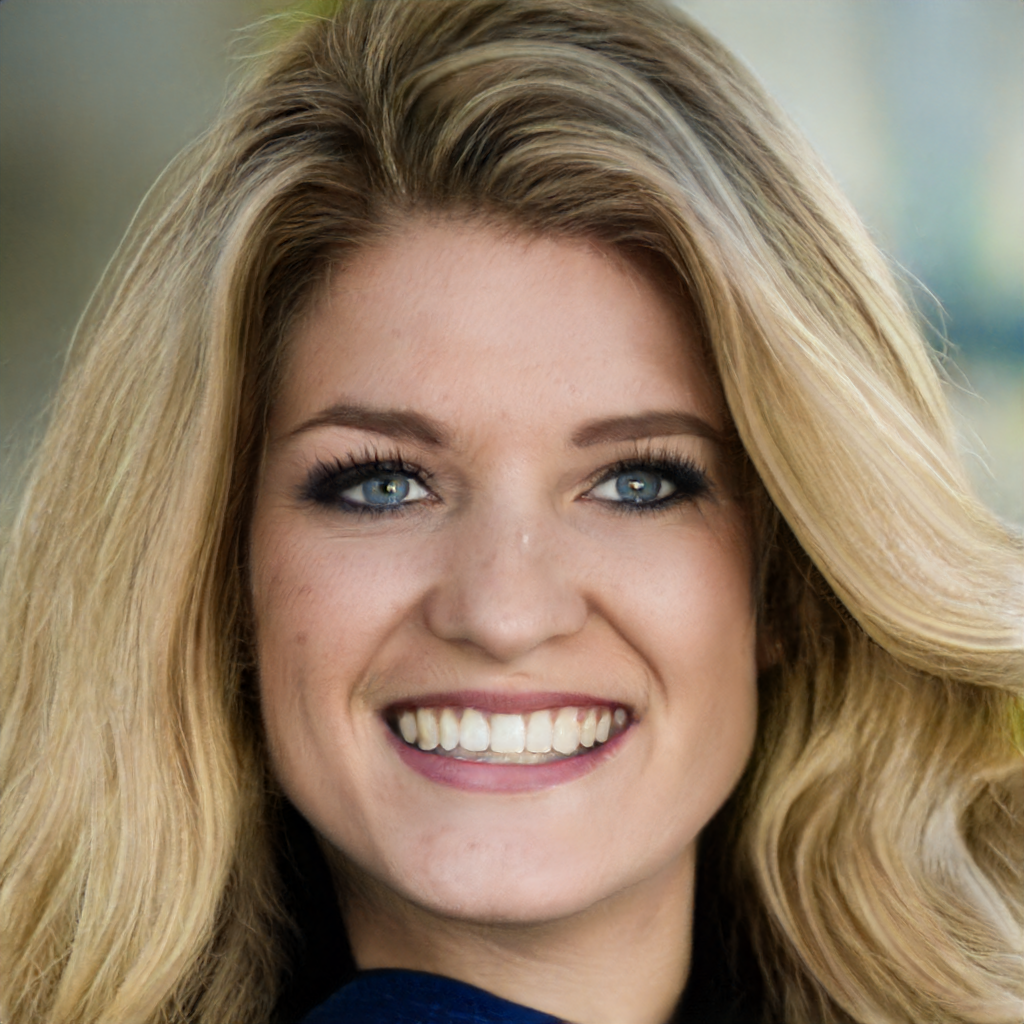}}
  \subfigure[\texttt{seed2048\_original}]{\includegraphics[width=0.48\textwidth]{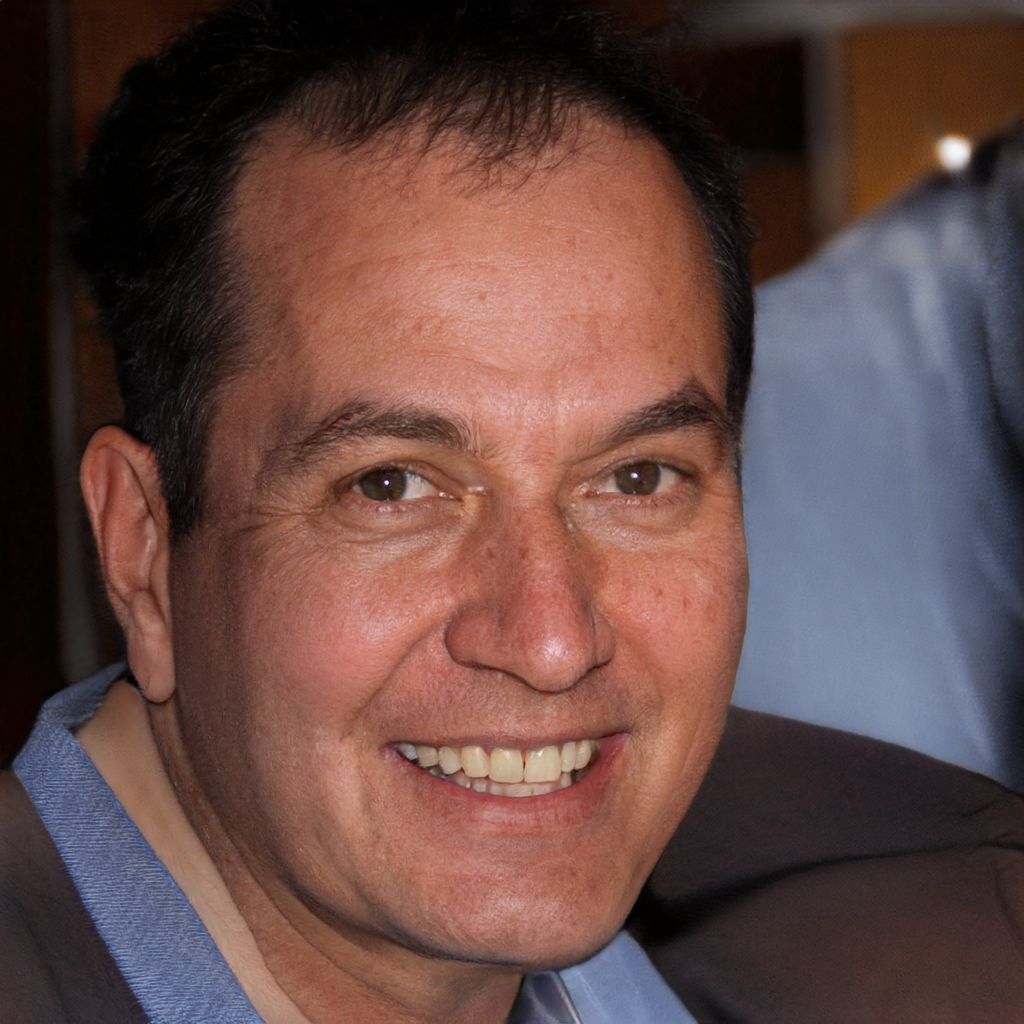}}
  \subfigure[\texttt{seed2048\_DGL-GAN}]{\includegraphics[width=0.48\textwidth]{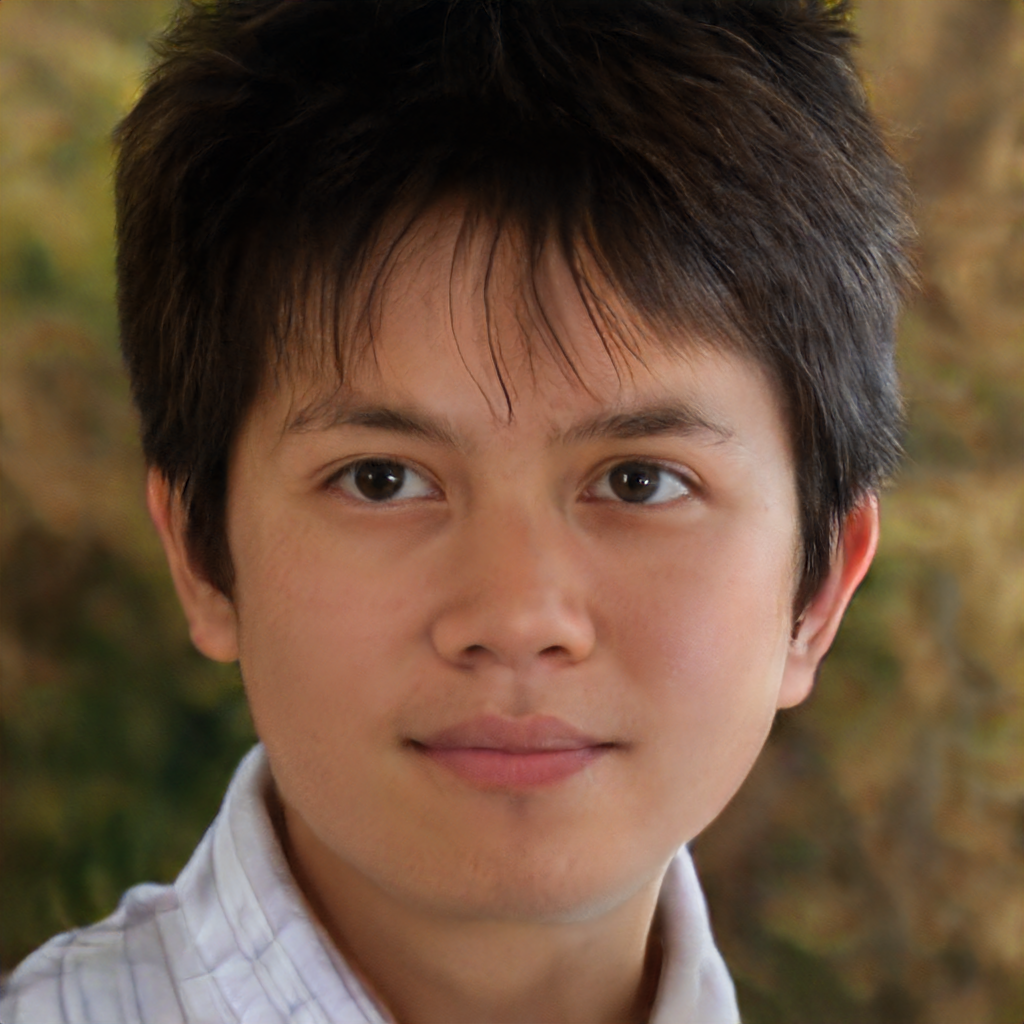}}
  \caption{Human faces generated by original StyleGAN2 and DGL-GAN with channel multiplier $1/2$. (a) and (c), (b) and (d) exploit the same $z$ under the same random seed.}
  \label{Fig: Generated faces}
\end{figure*}

\begin{figure*}[t]
    \vspace{-0.5cm}
    \centering
    \includegraphics[width=0.99\textwidth]{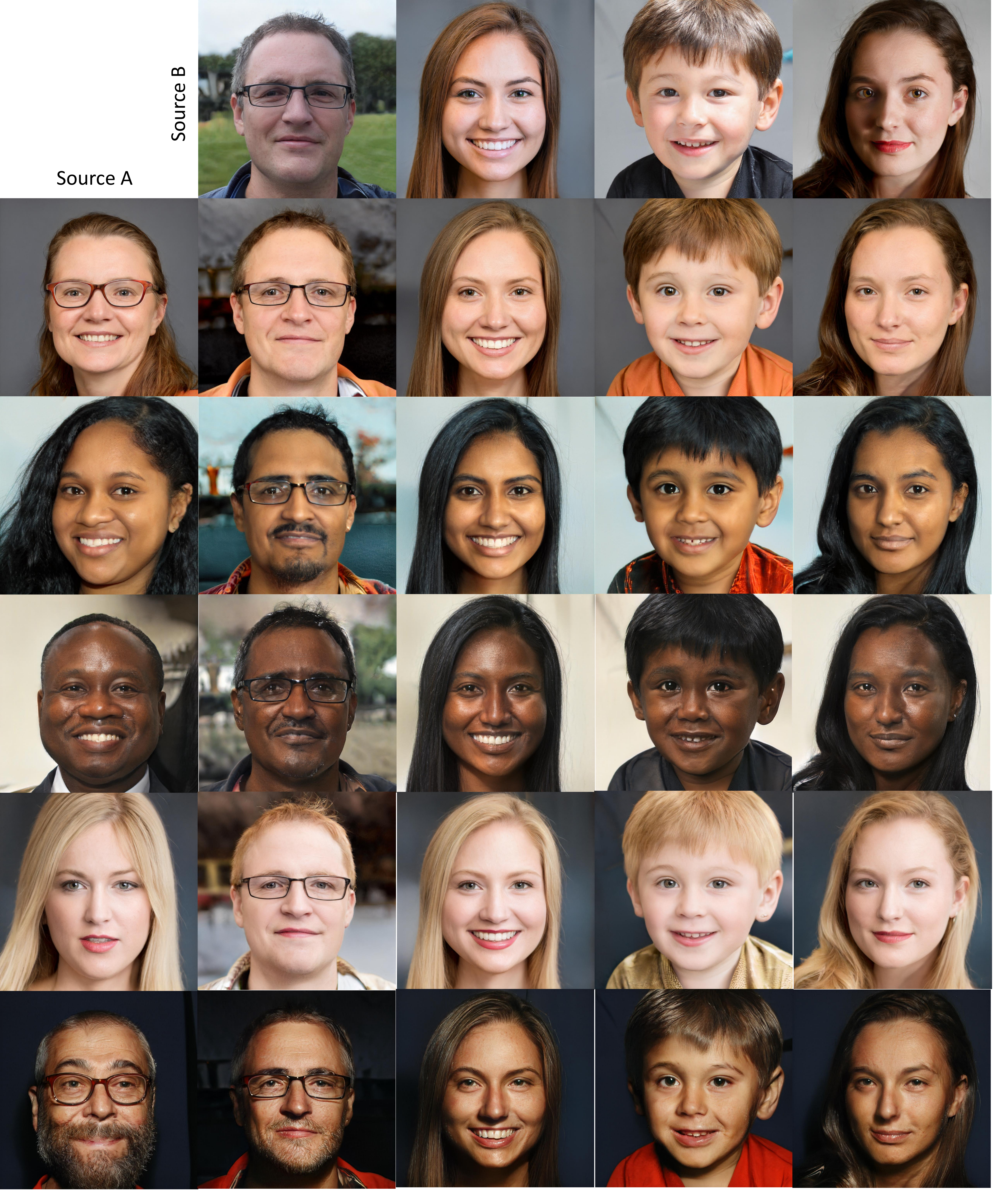}
    \caption{The generated images of original StyleGAN2. We obtain the images via mixing the intermediate latent $\mathcal{W}$ of two source images, i.e., Source A and Source B.}
    \label{Fig: StyleMixing_original}
\end{figure*}

\begin{figure*}[t]
    \vspace{-0.5cm}
    \centering
    \includegraphics[width=0.99\textwidth]{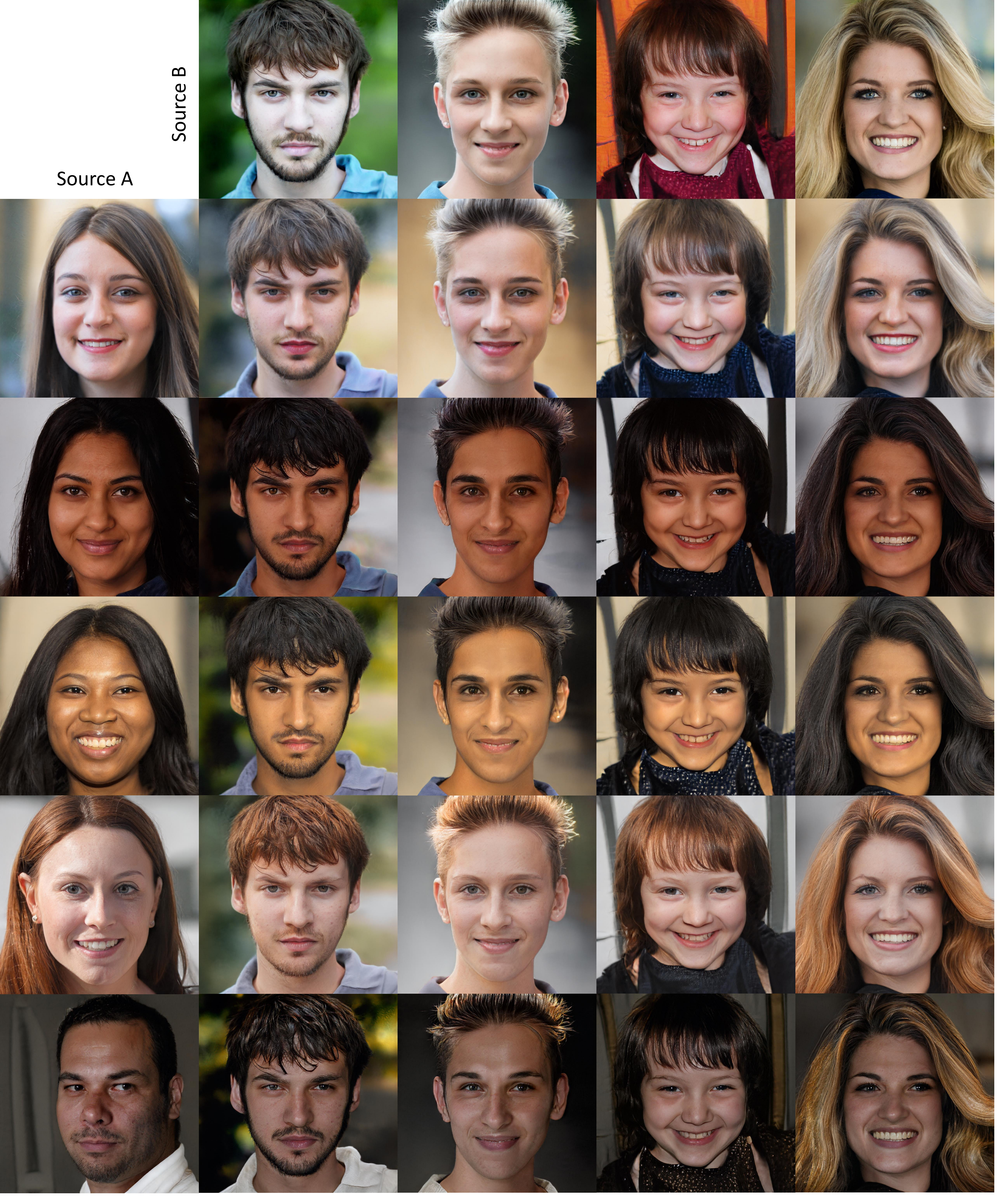}
    \caption{The generated images of DGL-GAN with channel multiplier $1/2$. We obtain the images via mixing the intermediate latent $\mathcal{W}$ of two source images, i.e., Source A and Source B.}
    \label{Fig: StyleMixing_1-2}
\end{figure*}

\begin{figure*}[t]
  \vspace{-0.5cm}
  \centering
  \subfigure[\texttt{InterpolateY\_orginal}]{\includegraphics[width=0.4\textwidth]{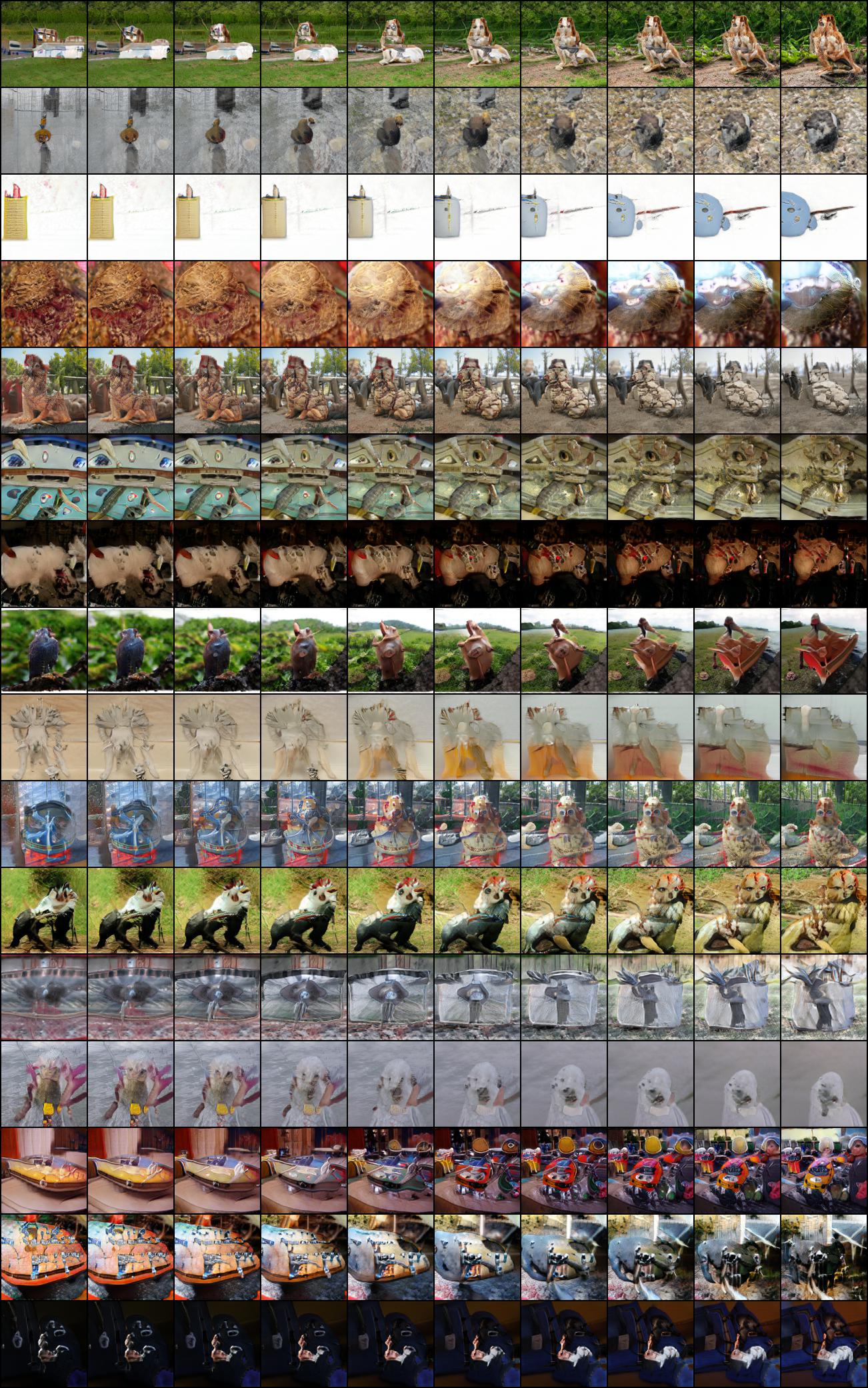}}
  \subfigure[\texttt{InterpolateY\_DGL-GAN}]{\includegraphics[width=0.4\textwidth]{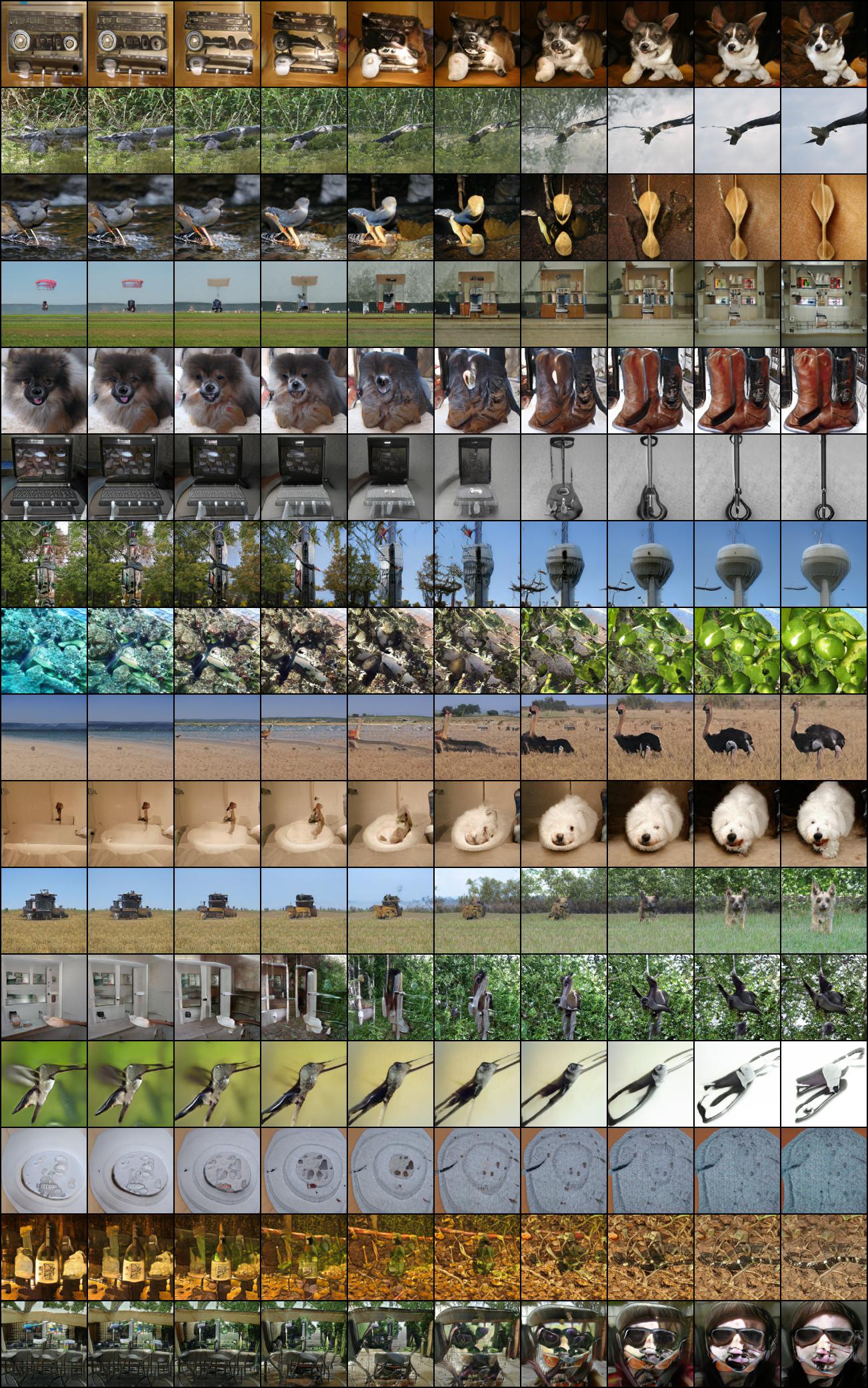}}
  \subfigure[\texttt{InterpolateZ\_original}]{\includegraphics[width=0.4\textwidth]{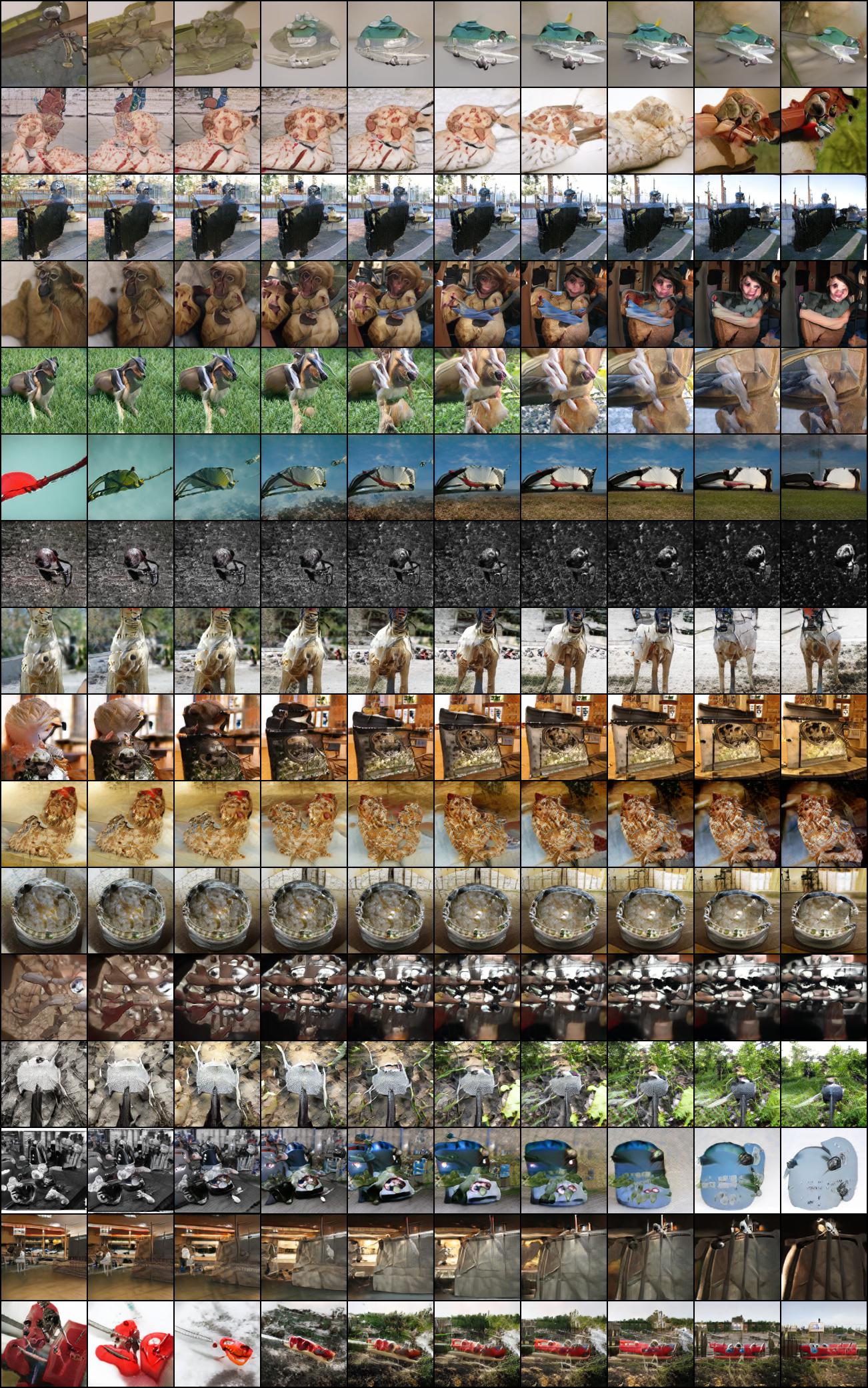}}
  \subfigure[\texttt{InterpolateZ\_DGL-GAN}]{\includegraphics[width=0.4\textwidth]{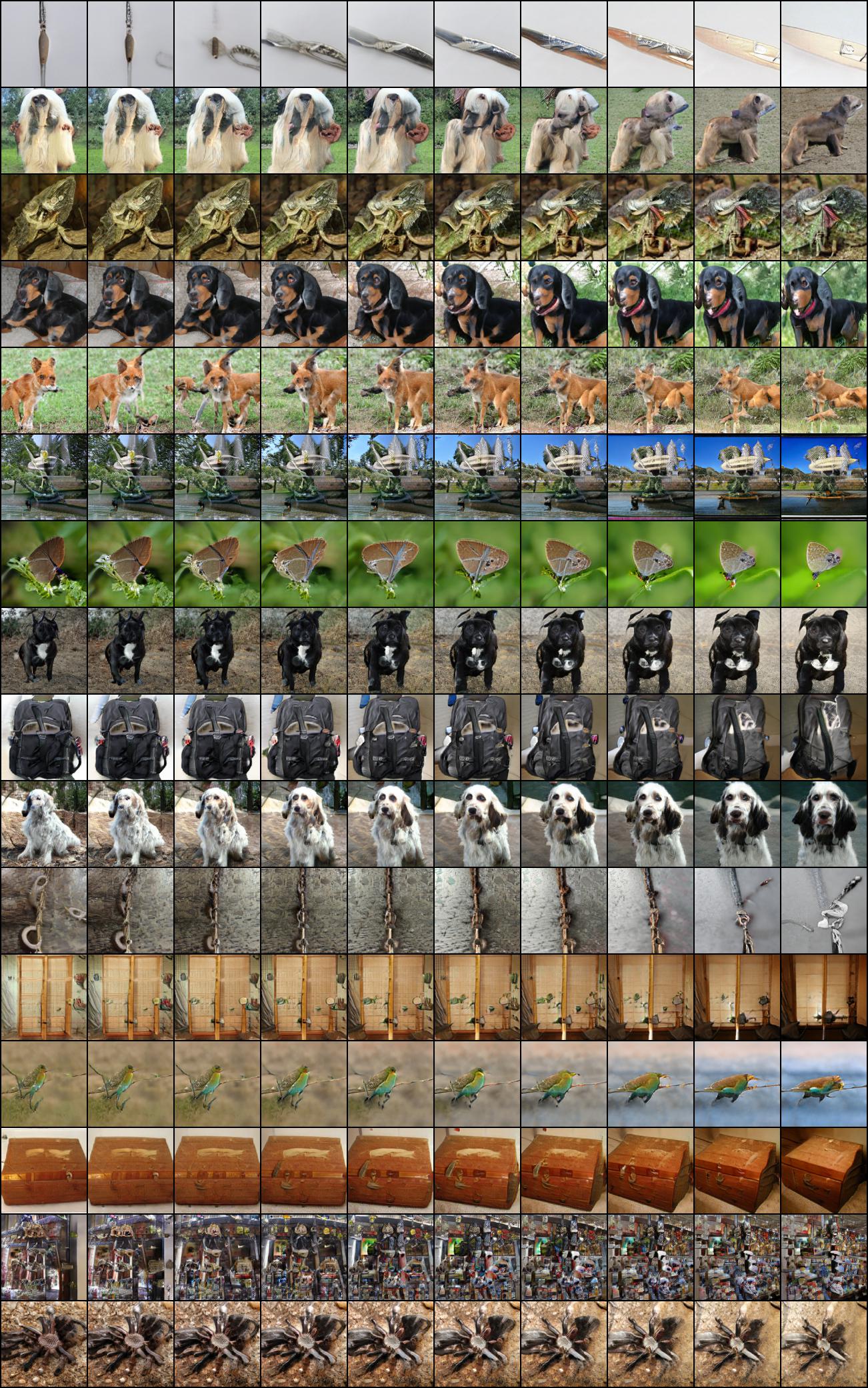}}
  \caption{The generated images of BigGAN and DGL-GAN with channel multiplier $1/2$, obtained via interpolating between ${\rm Y}_1$ and ${\rm Y}_2$ (label) or between ${\rm Z}_1$ and ${\rm Z}_2$ (noise).}
  \label{Fig: Generated ImageNet interpolation}
\end{figure*}

\section{Generated samples}
Figure \ref{Fig: Generated faces} shows the human faces with resolution $1024\times1024$ generated by original StyleGAN2 and DGL-GAN with channel multiplier $1/2$, respectively. With the same $z$, the generated faces of original StyleGAN2 and DGL-GAN share similar visual quality. We also utilize StyleGAN2 and DGL-GAN to generate style-mixing images, which are produced via interpolating between the intermediate latent $\mathcal{W}$ of two source images, shown in Figure \ref{Fig: StyleMixing_original} and Figure \ref{Fig: StyleMixing_1-2}, respectively. The style-mixing images of DGL-GAN show comparable visual quality with original StyleGAN2. Further, Figure \ref{Fig: Generated ImageNet interpolation} presents the interpolation images generated by original BigGAN and DGL-GAN with channel multiplier $1/2$, where the visual quality of images of BigGAN and DGL-GAN is similar, illustrating that DGL-GAN with the channel multiplier $1/2$ can even achieve the comparable performance from both quantitative perspective and qualitative perspective.




\end{appendices}


\bibliography{sn-bibliography}


\end{document}